\PassOptionsToPackage{hidelinks,breaklinks}{hyperref}

\documentclass{article}
\usepackage{arxiv}

\usepackage[utf8]{inputenc}
\usepackage[T1]{fontenc}
\usepackage{amsmath,amssymb,amsfonts}
\usepackage{booktabs}
\usepackage{array}
\usepackage[table]{xcolor}

\usepackage{graphicx}
\DeclareGraphicsExtensions{.pdf,.png,.jpg}
\usepackage{subcaption} 
\usepackage{nicefrac}
\usepackage{microtype}
\usepackage{textcomp}
\usepackage{url} 
\usepackage{tikz}
\usetikzlibrary{positioning,arrows.meta,shapes.multipart}
\usepackage{hyperref}
\hypersetup{
  colorlinks=true,
  linkcolor=brown,
  citecolor=green,
  urlcolor=blue
}

\title{Explaining Fine-Tuned LLMs via Counterfactuals: \\
A Knowledge Graph–Driven Framework}

\author{
  Yucheng Wang \quad Ziyang Chen \quad Md Faisal Kabir \thanks{Corresponding author. \\ 
  Accepted at KDD 2025 Structured Knowledge for Large Language Models Workshop (non-archival).} \\
  Penn State Harrisburg \\
  Pennsylvania, USA \\
  \texttt{\{yzw5780, zpc5231, mpk5904\}@psu.edu} \\
}

\begin{document}
\maketitle
\begin{abstract}
The widespread adoption of Low-Rank Adaptation (LoRA) has enabled large language models (LLMs) to acquire domain-specific knowledge with remarkable efficiency. However, understanding how such a fine-tuning mechanism alters a model's structural reasoning and semantic behavior remains an open challenge. This work introduces a novel framework that explains fine-tuned LLMs via counterfactuals grounded in knowledge graphs. Specifically, we construct BioToolKG, a domain-specific heterogeneous knowledge graph in bioinformatics tools and design a counterfactual-based fine-tuned LLMs explainer (\texttt{CFFTLLMExplainer}) that learns soft masks over graph nodes and edges to generate minimal structural perturbations that induce maximum semantic divergence. Our method jointly optimizes structural sparsity and semantic divergence while enforcing interpretability preserving constraints such as entropy regularization and edge smoothness. We apply this framework to a fine-tuned LLaMA-based LLM and reveal that counterfactual masking  exposes the model's structural dependencies and aligns with LoRA-induced parameter shifts. This work provides new insights into the internal mechanisms of fine-tuned LLMs and highlights counterfactual graphs as a potential tool for interpretable AI.
\end{abstract}

\section{Introduction}
In recent years, with the continuous advancement of NLP technologies and large language models (LLMs) \cite{mann2020language}, which exhibit remarkable generative capabilities \cite{mann2020language} and cross-domain generalization \cite{bommasani2021opportunities}, an increasing number of application-level techniques have begun to play a significant role across various domains \cite{hadi2023survey}.

Structured knowledge graphs \cite{fabian2007yago} have become a compelling direction in a wide range of Natural Language Processing(NLP) applications. As a form of heterogeneous information network (HIN), knowledge graphs have supported numerous early use cases, including recommendation systems, link prediction, and information fusion \cite{shi2016survey}. Recently, the integration of knowledge graphs with LLMs has given rise to novel applications such as domain-specific knowledge graph construction \cite{abu2021domain} and graph-enhanced retrieval-augmented generation (RAG) \cite{gao2023retrieval}, significantly improving the factuality and controllability of LLM outputs. However, due to the low information purity and the presence of substantial irrelevant content in many knowledge graphs, using RAG may lead LLMs to hallucinate or fail to extract accurate subgraphs as reliable external knowledge \cite{agrawal2023can, wu2023task}. Furthermore, the extracted subgraphs often contain cycles, which can pose significant challenges in tasks such as knowledge path reasoning and graph ordering \cite{wallace2024pruning}.

Meanwhile, with the rise of explainable AI (XAI) techniques, there has been growing attention to the user-friendliness and interpretability of AI systems in real-world applications \cite{ribeiro2016should}. A variety of explainability frameworks have been proposed for predictive models, utilizing diverse strategies. Among these, many rely on perturbing input features to evaluate their influence on model outputs \cite{lundberg2017unified, ribeiro2016should, mothilal2020explaining, ribeiro2018anchors, guidotti2018local}. However, it is important to note that alternative approaches—such as gradient-based methods \cite{sundararajan2017axiomatic} or attention-based \cite{vaswani2017attention} explanations—also exist. However, there are additional challenges in interpreting LLMs. Firstly, the output of an LLM is not a simple classification result, but rather contains rich semantic content. Secondly, due to the complexity of LLM architectures and the massive scale of training data, fine-grained explanations based on attention mechanisms alone are often insufficient or intractable \cite{jain2019attention}. Finally, prompt-based self-explanation tools for LLMs have shown limited effectiveness, primarily due to the prevalence of hallucinations and the inherent uncertainty in LLM outputs \cite{turpin2023language, huang2023can}.

To improve and extend existing methods, this paper proposes the following innovations in the subsequent sections:
\begin{enumerate}
    \item We introduce \textbf{BioToolKG}, a domain-specific and semantically structured knowledge graph that organizes bioinformatics tools, algorithms, databases, and more related entities;

    \item A novel \textbf{counterfactual-based interpretability framework} (\texttt{CFFTLLMExplainer}) is proposed, specifically tailored for fine-tuned large language models, enabling structure-aware explanation of model behavior;

    \item The counterfactual generation is formalized as an \textbf{unsupervised optimization problem}, where a multi-objective loss jointly balances semantic divergence, structural sparsity, prompt relevance, and graph smoothness;

    \item To uncover the internal decision mechanisms of fine-tuned LLMs, we conducted a \textbf{multi-perspective interpretability analysis} by aligning learned structural masks, token-level attention scores, and LoRA-induced embedding shifts.
\end{enumerate}

Following the introduction, the paper is organized into several sections. Section~\ref{sec:related} reviews relevant works on LLM interpretability and outlines the motivation for our study. Section~\ref{sec:method} introduces our proposed counterfactual-based interpretability framework tailored for fine-tuned LLMs. Section~\ref{sec:experiments} introduces experiments for revealing the mechanisms in LoRA fine-tuning process. In Section~\ref{sec:result} presents preliminary experimental results that demonstrate the effectiveness of our approach. Finally, Section~\ref{sec:future} and Section~\ref{sec:conclusion} discuss potential directions for future work and summarize main contributions.

\section{Related Works and Motivations} \label{sec:related}

With the rapid development of large language models (LLMs), the widespread adoption of Low-Rank Adaptation (LoRA) fine-tuning has played a significant role in enabling LLMs to acquire domain-specific knowledge efficiently, thereby facilitating the construction of expert systems. This technique is particularly important for the broader deployment of LLMs in specialized applications. Specifically, LoRA introduces a low-rank matrix as an incremental module into the existing model architecture, effectively altering the model’s behavior. By inserting these trained adapter layers into the original model, LoRA enables the acquisition of new knowledge while updating only a small subset of parameters, thus ensuring efficiency and adaptability \cite{hu2022lora}.

With the rapid advancement of Explainable AI (XAI), a variety of XAI tools, such as SHAP, LIME and DiCE, have been widely applied to interpret and analyze different AI models and systems. In LIME, feature importance is indicated by the weights assigned under feature perturbation \cite{ribeiro2016should}, and in SHAP, it is represented by computing Shapley values \cite{lundberg2017unified}. Under this background, the interpretability of large language models (LLMs) has increasingly attracted widespread attention alongside their rapid development \cite{zhao2024explainability}. Meanwhile, counterfactual explanations—such as those generated by DiCE—enhance model interpretability and human alignment by producing diverse and informative counterfactuals. Specifically, DiCE formulates counterfactual generation as a multi-objective optimization problem, where a carefully designed loss function minimizes the distance between the counterfactuals and the original instances while simultaneously altering the model's prediction \cite{mothilal2020explaining}.

TokenSHAP, tries to explain LLM output from token level. Specifically, it extends the application of SHAP by introducing Shapley value-based attribution to natural language processing tasks, enabling a deeper understanding of how different components of an input prompt influence the model’s output. By incorporating Monte Carlo Shapley Estimation, it achieves a balance between computational efficiency and estimation accuracy \cite{horovicz2024tokenshap}.

Some recent studies have proposed embedding interpretability frameworks directly into the architecture of large language models (LLMs) to overcome the limitations of traditional black-box interpretability, which relies solely on analyzing model outputs. Concept Bottleneck Large Language Models (CB-LLMs) introduce an inherently interpretable framework for LLMs, demonstrating clear advantages in terms of scalability and transparency—both critical for the responsible development. Specifically, the core idea of CB-LLMs lies in training a Concept Bottleneck Layer (CBL) that maximizes the concept score (i.e., similarity) of input samples, thereby generating transparent concept weights that explain the model’s outputs in a semantically meaningful and interpretable manner \cite{sun2024concept}.

Overall, current research on the interpretability of large language models (LLMs) generally follows two main tracks. The first focuses on analyzing attention weights to investigate whether they capture causal relationships between input tokens \cite{vaswani2017attention}. However, attention-based explanations are often unreliable proxies for model reasoning, particularly in large-scale models where deep, layered architectures and long-range context dependencies complicate attribution \cite{jain2019attention, wiegreffe2019attention, serrano2019attention}. The second line of work centers on prompt engineering, where modifications to the input prompt are used to probe model behavior \cite{webson2022prompt, zhou2022least}. While effective for black-box models, such approaches lack a principled connection to internal mechanisms and are sensitive to prompt design, contextual variation, and output stochasticity \cite{pasquini2024neural, zhao2021calibrate}. Some studies adapt classical interpretability tools, such as SHAP and Integrated Gradients, to LLMs \cite{horovicz2024tokenshap, enguehard2023sequential}. Yet these methods, originally designed for classification tasks, are often ill-suited for generation settings, where token interactions are complex and the output is inherently structured as free-form text \cite{zhao2024explainability}. Some recent work has also explored self-explanation approaches, prompting LLMs to generate natural language rationales. While this improves human interpretability, such explanations often lack verifiability and alignment with the model’s internal decision process \cite{zhao2024explainability, huang2023can, xu2024sayself}.

Intuitively, one might hope to leverage prompt-based methods to infer the relevance of specific attention weights by analyzing model outputs, thereby facilitating rapid identification of the components most responsible for a given prediction. However, this approach faces several major challenges. First, LLM outputs are typically general, semantically rich, and unstructured, which raises the question of how to encode such free-form text into formats that are trainable and operable, such as structured representations or symbolic features. Second, under this framework, we implicitly assume that the output can be meaningfully aligned with internal attention weights. Yet in practice, establishing such alignment is nearly infeasible given the enormous number of parameters and the complexity of attention pathways in modern LLMs.

\section{Methodology} \label{sec:method}
This section introduces a counterfactual-based interpretability framework for fine-tuned LLMs. It covers the problem statement (Section.\ref{subsec:ps}), the construction and injection of BioToolKG (Section.\ref{subsec:kgconstrction}) and our method \texttt{CFFTLLMExplainer} for explaining fine-tuned LLMs (Section.\ref{subsec:framework}).

Specifically, a structured knowledge graph is considered a heterogeneous graph, which facilitates knowledge injection and knowledge reconstruction from LoRA fine-tuned LLM. After the injection, we offer an unsupervised learning method for generating counterfactuals by learning trainable soft masks on the structured knowledge representation. 

\subsection{Problem Statement} \label{subsec:ps}
In this paper, a Knowledge Graph is defined as a heterogeneous information network \cite{shi2016survey} \cite{sun2013mining, sun2009ranking}, as\\
$G = (V, E, \mathcal{A}, \mathcal{R}, \varphi, \psi, X^V, X^E)$, where $V$ is a finite set of nodes, $E$ is a finite set of edges, $\varphi : V \rightarrow \mathcal{A}$ is an object (entities) mapping function, $\psi : E \rightarrow \mathcal{R}$ is a link (edge) type mapping function, $X^V = \{x_v | v \in V\}$ and each $x_{v}$ is the attribute vector of node,  $X^E = \{x_e | e \in E\}$ and each $x_e$ is the attribute vector of edge, $|\mathcal{A}| > 1$ and $|\mathcal{R}| > 1$. 

Each object $v \in V$ belongs to one particular object type in the object type set $\mathcal{A} : \varphi(v) \in \mathcal{A}$, and each link $e \in E$ belongs to a particular relation type in the relationship type set $\mathcal{R} : \psi(e) \in \mathcal{R}$. 

A Counterfactual sample (CF sample) $\bar{x}$ for an instance $x$ according to a trained classifier $f$ is found by perturbing the features of $x$ such that $f(x) \neq f(\bar{x})$ \cite{wachter2017counterfactual}. An Optimal CF sample $\bar{x}^*$ is one the minimizes distance between the original instance and the CF sample, according to some distance function $d$, and the resulting optimal CF explanation is $\Delta_x^* = \bar{x}^* - x$ \cite{lucic2022focus}.

We consider a LoRA-adapted transformer-based large language model, where the original weight matrix $W_0 \in \mathbb{R}^{d \times k}$ in selected modules (e.g., attention projections) is frozen, and a low-rank residual update is learned via trainable adapter matrices $A \in \mathbb{R}^{d \times r}$ and $B \in \mathbb{R}^{r \times k}$, where $r \ll \min(d, k)$. The adapted weight is defined as:
\begin{equation}
W = W_0 + \Delta W = W_0 + \alpha \cdot AB
\end{equation}

, where $\alpha$ is a scalar scaling factor controlling the update magnitude. During fine-tuning, only the adapter parameters $A$ and $B$ are updated, while $W_0$ remains fixed \cite{hu2022lora}. In this setting, the LoRA adapter effectively defines a low-rank subspace $\mathrm{span}(A)$ in which the model's behavior is modulated. Our objective is to investigate whether and how these learned subspaces encode structure-sensitive patterns when the model is presented with knowledge graph inputs.


\subsection{BioToolKG Construction and Injection} \label{subsec:kgconstrction}
\begin{figure}[htbp]
    \centering
    \includegraphics[width=0.6\columnwidth]{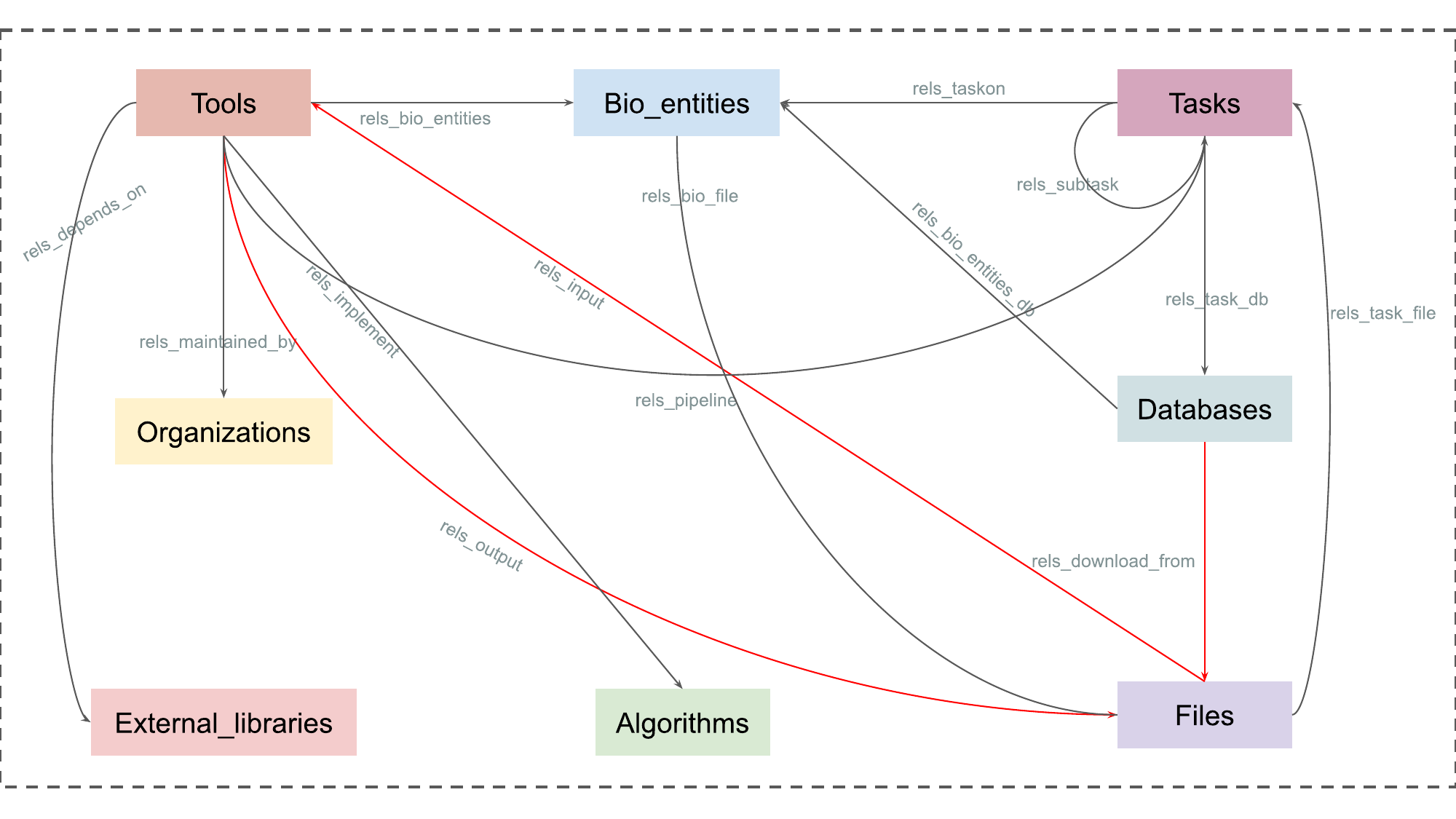}
    \caption{Definitions of entities and relations in BioToolKG}
    \label{fig:kg}
\end{figure}

Figure.\ref{fig:kg} provides a comprehensive overview of the entities and relationships encoded within BioToolKG. The knowledge graph comprises eight distinct entity types and fourteen relation types, reflecting the complexity and richness of the bioinformatics domain. Although knowledge graphs can support a wide range of downstream tasks, this study just focuses on the tool usage Pathfinding problem. Typically started from a database entity. Within this context, concentrating on three core relation types : \\
``\texttt{rels\_download\_from}'', ``\texttt{rels\_input}'', and ``\texttt{rels\_output}'', which are critical for modeling data flow across tools. These relations are visually distinguished as red-directed edges in Figure.\ref{fig:kg} to emphasize their role in enabling pipeline construction. The tool usage Pathfinding problem can be defined as follows :

The goal of the Pathfinding task is to construct a valid tool execution pipeline $P = [v_1, v_2, \dots, v_n]$ satisfying the following constraints:

\begin{itemize}
    \item $v_1$ is a database-type node that serves as the starting input;
    \item Every tool node $v_i$ in the path must consume one or more input files from a preceding file node $f_{i-1}$ such that $(f_{i-1}, v_i) \in E$ and $\psi(f_{i-1}, v_i) = \text{input}$;
    \item Each tool node $v_i$ must produce one or more output files $f_i$ such that $(v_i, f_i) \in E$ and $\psi(v_i, f_i) = \text{output}$;
    \item The output file(s) $f_i$ of tool $v_i$ must be consumable by the next tool $v_{i+1}$, i.e., $(f_i, v_{i+1}) \in E$ and $\psi(f_i, v_{i+1}) = \text{input}$;
    \item The path terminates at a tool node that fulfills the task goal (in this paper, a virtual \texttt{Evaluation Information} node is utilized for termination).
\end{itemize}

The interpretability of LLMs is inherently complicated. This paper specifically focuses on the explainability of fine-tuned LLMs by analyzing the differences in model outcomes before and after fine-tuning, trying to uncover insights into the internal mechanisms of LLMs fine-tuning process. Although knowledge injection plays a necessary role in our framework, we do not center our attention on the quality of knowledge injection evaluation. Instead, we concentrate on interpreting the impacts of LoRA-tuned adapters on the behavior of the model.

In this study, the baseline model, ``DeepSeek-R1-Distill-Llama-8B'', utilizes the ``LLaMA-8B'' architecture and undergoes supervised fine-tuning on instruction-style reasoning data generated by ``DeepSeek-R1''. Through this teacher–student (distillation) process, ``DeepSeek-R1''’s reasoning capabilities are transferred to a smaller, more efficient student model(``LLaMA-8B'') \cite{deepseek2024llama8b, deepseekai2025deepseekr1incentivizingreasoningcapability}. In our method, an instruction-style format is adopted to construct the fine-tuning dataset for knowledge injection. Each instance is constructed as an prompt-response pair, which aligns with standard LoRA/PEFT fine-tuning framework.

\subsection{Counterfactual-based Fine-tuned LLMs Explainer (\texttt{CFFTLLMExplainer})} \label{subsec:framework}
In this section, we propose an unsupervised machine learning framework for generating optimal counterfactual samples on heterogeneous graphs and show how counterfactuals can help to interpret the internal mechanisms of fine-tuned LLMs.

One of the key design goals is to achieve significant semantic divergence in the model's output through minimal and interpretable structural modifications to the input graph. A traditional counterfactual example $\bar{x}$ for an instance $x$ is defined as a perturbed version such that a trained classifier $f$ changes its prediction: $f(\bar{x}) \neq f(x)$ \cite{wachter2017counterfactual}. The optimal counterfactual example $\bar{x}^*$ is the one that minimizes the distance to the original input according to some distance function $d(\cdot, \cdot)$:
\begin{equation}
    \bar{x}^* = \arg\min_{\bar{x}} \left\{ d(x, \bar{x}) \ , \ \text{s.t.} \ f(\bar{x}) \neq f(x) \right\}
\end{equation}
The difference $\Delta_x^* = \bar{x}^* - x$ constitutes the counterfactual explanation \cite{lucic2022focus}. In the context of graphs, particularly heterogeneous knowledge graphs (KGs), we redefine counterfactual examples as perturbed subgraphs $G_c$ that cause a semantic shift while retaining minimal structural changes. To formalize the generation of optimal KG counterfactuals, we propose a multi-objective loss function, can be simplified represented as follows:
\begin{equation}
    \mathcal{L} = \mathcal{L}_{structure} + \alpha \cdot \mathcal{L}_{semantic\_divergence}
\end{equation}
, where $\mathcal{L}_{structure}$ constrains structural perturbations by assigning higher costs to extensive edge or node modifications, promoting sparsity. In contrast, $\mathcal{L}_{semantic\_divergence}$ encourages producing divergent semantic interpretations between the original and counterfactual graphs. 

Each node $v \in V$ is assigned a learnable mask $\mathbf{m}_v \in [0,1]$, and each edge $e \in E$ is assigned a learnable mask $\mathbf{m}_e \in [0,1]$. Mask values are sampled using the Gumbel-Sigmoid reparameterization for differentiability during training. Gumbel-Sigmoid represents a differentiable approximation to sampling, wherein continuous variables are used to approximate binary (e.g., Bernoulli) sampling, thereby allowing gradient-based optimization via backpropagation. Formally, 
\begin{equation}
    \tilde{z} = \sigma\left( \frac{log \ \alpha + G}{\tau} \right)
\end{equation}
, where $G = -log(-log(U + \varepsilon) + \varepsilon), \ U \sim \text{Uniform}(0, 1)
$ is the noise, $\tilde{z}$ each differentiable mask, $\sigma(x) = \frac{1}{1 + e^{-x}}$ is the sigmoid function, and $\tau$ is the temperature. 

The original graph is converted to a textual format $\mathcal{T}(G)$ using node attributes and edge relations. The masked graph $G_c$ is similarly converted to $\mathcal{T}(G_c)$ based on active nodes/edges. Semantic loss is then computed between $\mathcal{T}(G)$ and $\mathcal{T}(G_c)$.

To ensure the factual correctness and structural interpretability, we enforce multiple constraints and penalties, including entropy regularization, sparsity control, and edge smoothness regularization based on node importance (similarity with prompts).

The overall loss $\mathcal{L}_{\text{total}}$ is composed of the following components:
\begin{align} \label{eq:overall}
\mathcal{L}_{\text{total}} = \;
& \mathcal{L}_{\text{structure}} +
\alpha \cdot \mathcal{L}_{\text{semantic}} +
\beta \cdot \mathcal{L}_{\text{entropy}} \notag \\ 
&+
\gamma \cdot \mathcal{L}_{\text{preserve}} +
\delta \cdot \mathcal{L}_{\text{hard}} +
\epsilon \cdot \mathcal{L}_{\text{smooth}}
\end{align}

Each term is defined as follows:

Structure sparsity loss:
\begin{equation}
\mathcal{L}_{\text{structure}} =
\lambda_V \cdot \sum_{v \in V} \lambda_v \cdot m_v +
\lambda_E \cdot \sum_{e \in E} \lambda_{\psi(e)} \cdot m_e
\end{equation}
, where $\lambda_v$ and $\lambda_{\psi(e)}$ denote the regularization weights of node $v$ and edge $e$ based on their types.

Semantic loss:
\begin{equation}
\mathcal{L}_{\text{semantic}} =
1 - \cos \left( \text{TF-IDF}(\mathcal{T}(G)),\ \text{TF-IDF}(\mathcal{T}(G')) \right)
\end{equation}
, where cos($\cdot$) is cosine similarity. In this work, TF-IDF is adopted as the embedding strategy due to the observation that pretrained embedding models, while exhibiting powerful semantic comprehension, tend to exhibit low sensitivity to the nuanced semantics of pipeline structures. 
  
Optionally reweighted by prompt relevance:
\begin{align}
\mathcal{L}_{\text{semantic}} 
&\leftarrow w_{\text{prompt}} \cdot \mathcal{L}_{\text{semantic}}, \notag \\
w_{\text{prompt}} 
&= 1 + \sum_{v \in V} (1 - m_v) \cdot \text{sim}(x_v, \ \texttt{prompt})
\end{align}
, where sim($\cdot$) is also the combination of TF-IDF and cosine similarity. By dynamically assigning importance to nodes, the model increases the semantic discrepancy when it removes nodes with high similarity to the prompt. This mechanism is particularly important in pipeline graphs, which typically contain a limited number of nodes and edges, in order to prevent degenerate behavior, such as the model removing all nodes simply to maximize semantic divergence.

Entropy regularization : similarly, due to the limited scale of the graph, the model may struggle to converge in certain cases. To address this, entropy regularization is applied to encourage the node and edge masks to approach binary values (i.e., closer to 0 or 1).
\begin{align}
\mathcal{L}_{\text{entropy}} =\;
& - \sum_{v \in V} \left[ m_v \log m_v + (1 - m_v) \log(1 - m_v) \right] \notag \\
& - \sum_{e \in E} \left[ m_e \log m_e + (1 - m_e) \log(1 - m_e) \right]
\end{align}

Minimum structure preserve loss (structural stability constraint):
\begin{equation}
\mathcal{L}_{\text{preserve}} =
\text{ReLU} \left( |\{ v \in V \mid \varphi(v) = \texttt{Tool} \}| - 1 - \sum_{e \in E}
\mathbf{1}[m_e \ge 0.5] \right)
\end{equation}

Hard mask retention penalty (structural stability constraint):
\begin{equation}
\mathcal{L}_{\text{hard}} =
\sum_{e \in E^{\text{hard}}} \text{ReLU}(0.5 - m_e) +
\sum_{v \in V^{\text{hard}}} \text{ReLU}(0.5 - m_v)
\end{equation}

Edge-mask smoothness regularization (based on node importance weights): since our task is grounded in realistic bioinformatics network applications, we aim to maximize semantic discrepancy while avoiding completely isolated edges and nodes. To this end, we introduce smoothness regularization to smooth the learned weights, thereby enhancing the interpretability and semantic coherence of the resulting graph $G_c$.
\begin{equation}
\mathcal{L}_{\text{smooth}} =
\frac{1}{|E|} \sum_{e = (u, v) \in E}
\left( m_e - \frac{w_{\varphi(u)} \cdot m_u + w_{\varphi(v)} \cdot m_v}{w_{\varphi(u)} + w_{\varphi(v)}} \right)^2
\end{equation}
, where $w_{\varphi(u)}$ and $w_{\varphi(v)}$ denote the weight of node types.

\section{Experiments Design}
\label{sec:experiments}

Two experiments are designed to investigate how \texttt{CFFTLLMExplainer} can be used to interpret LoRA fine-tuned LLMs. Our goal is to understand whether structural information from a knowledge graph is preserved or transformed during fine-tuning, and to identify the key structural components that drive semantic changes in the model's output.

\subsection{Experiment 1: Counterfactual-based Interpretation} \label{experiment1}
A set of preliminary experiments is conducted to investigate two central questions: (1) whether the LoRA-fine-tuned LLM exhibits structural dependencies on specific nodes or edges with the structured input, and (2) whether such dependencies are reflected in the learned parameters of the LoRA adapters.

\begin{figure}[htbp]
    \centering
    \includegraphics[width=0.6\columnwidth]{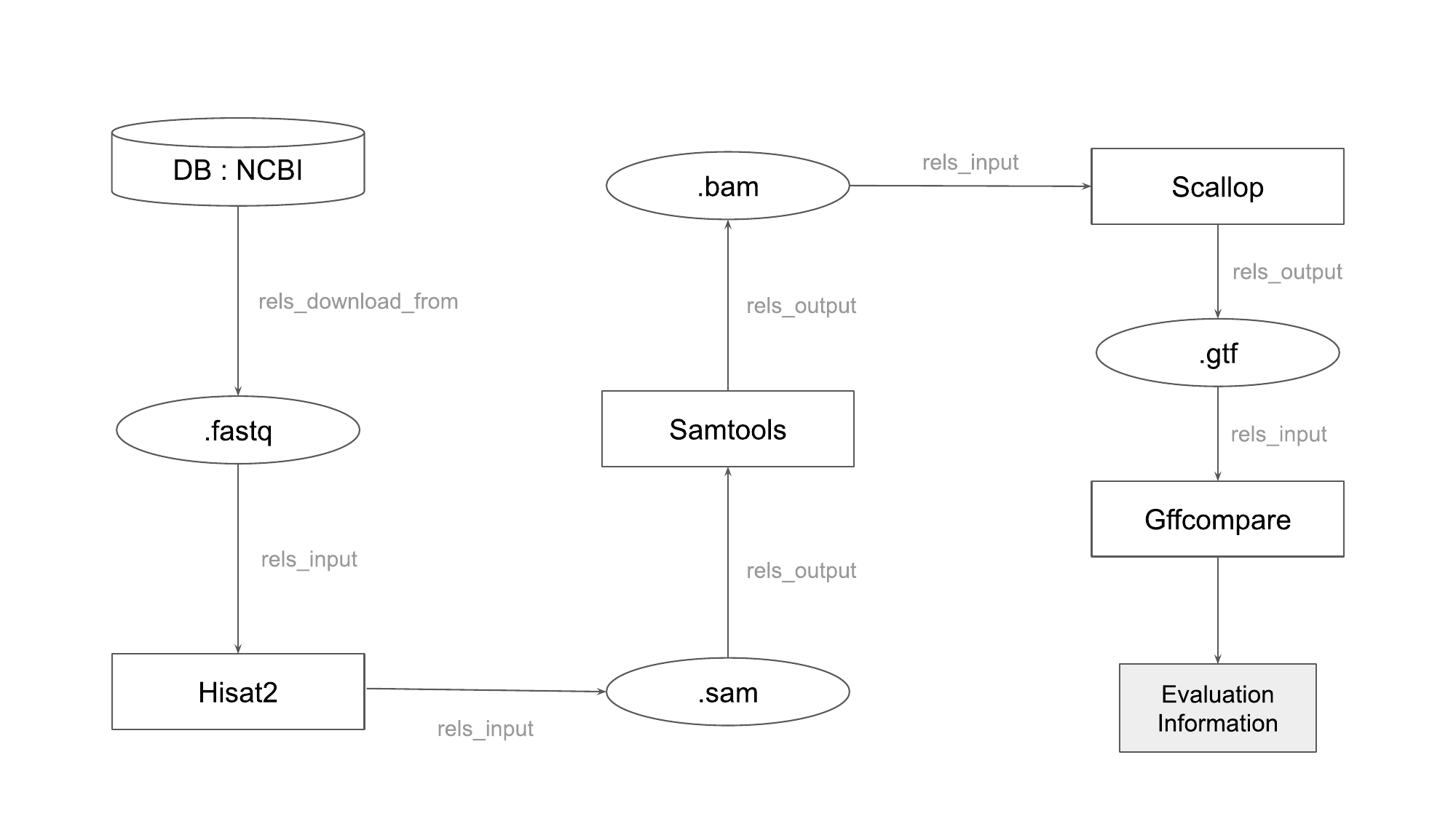}
    \caption{One Transcript Assembly pipeline}
    \label{fig:scallop}
\end{figure}

A biological toolchain graph $G$ (Figure~\ref{fig:scallop}) is first constructed to satisfy the requirements of a Pathfinding task (Section~\ref{subsec:kgconstrction}). A counterfactual subgraph $G_c$ is then derived through \texttt{CFFTLLMExplainer} (Section~\ref{subsec:framework}). The resulting subgraph $G_c$ is optimized to induce maximal semantic deviation while preserving minimal structural changes, thereby enabling the examination of whether the LLM exhibits structural dependencies on the input graph. Both $G$ and $G_c$ are subsequently converted into textual prompts using the same prompt template, and are provided as inputs to both the baseline and fine-tuned models for comparison. It is notable that in \texttt{CFFTLLMExplainer}, the counterfactual model is trained independently of the LLM, using TF-IDF similarity as a surrogate signal. This disentanglement enables generalizable and interpretable counterfactual explanations of fine-tuned LLM behavior. This enables subsequent analysis from multiple perspectives as follows:

Assume that baseline model $f_{base}$ and fine-tuned model $f_{ft}$, bioinformatics toolchain graph $G$ and counterfactual sample $G_c$. 
\begin{itemize}
    \item \textit{Semantic Drift Analysis}. Bioinformatics toolchains are extracted from $f_{\text{base}}(G)$, $f_{\text{ft}}(G)$, and $f_{\text{ft}}(G_c)$. To quantify the semantic shifts induced by structural perturbations, several metrics are computed: (1) Jaccard similarity $J(A, B) = \frac{|A \cap B|}{|A \cup B|}$, (2) edit distance $d_{\text{edit}}(A, B)$ measuring the minimum number of insertions, deletions, or substitutions to convert sequence $A$ to $B$, (3) path overlap defined as $P(A, B) = \frac{|\text{prefix}(A, B)|}{\min(|A|, |B|)}$ where $\text{prefix}(A, B)$ denotes the length of the longest common prefix;
    \item \textit{Attention Alignment Evaluation}. The goal is to extract the average attention assigned to each structural token in $\mathcal{T}(G)$, with a particular focus on tool-type nodes. Since the nodes preserved in the counterfactual graph $G_c$ are typically less semantically influential, this experiment provides evidence that the structural mask tends to retain nodes with relatively low impact on the model’s semantic output;
    \item \textit{Adapter Shift Probing}. The latent shift is computed by projecting the token embeddings through the LoRA projection matrices $(B \cdot A)$. To investigate whether the LoRA fine-tuning process internally encodes structural preferences for key bioinformatics components, we probe the learned adapter parameters by measuring the latent representation shift introduced to each token embedding. Specifically, for a given node token embedding $e$, we compute the LoRA-induced directional shift $\Delta$ as: $\Delta = \mathbf{B} \cdot \mathbf{A} \cdot e$, where $\mathbf{A}$ and $\mathbf{B}$ denote the low-rank weight matrices of the LoRA adapter within the query projection module. The resulting vector $\Delta$ captures the fine-tuning-induced modulation of the token representation, and its $\ell_2$ norm serves as a proxy for the token's sensitivity under task-specific adaptation. If the masked nodes correspond to large latent shifts, it indicates that these nodes play a more critical role in the LoRA fine-tuning process. Consequently, masking such nodes poses a substantial challenge to the fine-tuned model, revealing their importance in the model's adapted behavior.
\end{itemize}

\subsection{Experiment 2: Baseline Perturbation} \label{baseline}
Although Experiment 1 provides preliminary evidence that the proposed counterfactual-based framework for explaining fine-tuned LLMs is effective, concerns remain regarding the framework’s overall rationality and advancement. Specifically, the following critical questions arise: “If we do not explicitly consider graph structure and semantic features, can random perturbations produce controllable semantic shifts?”, “Compared to random perturbations, does the \texttt{CFFTLLMExplainer} better preserve structural coherence while inducing semantic shifts?”, and “Can attention-based heuristic masks induce sufficient semantic differences?”. To address these questions, a series of baseline experiments are designed in this section as follows:
\begin{itemize}
    \item Random nodes perturbation: random fixed number of nodes
    (same as $G_c$) and relative edges are masked, denoted by \texttt{Randomnodemask}; 
    \item Random edges perturbation: random fixed number of edges(same as $G_c$) are masked, denoted by \texttt{Randomedgemask};
    \item Random nodes and edges perturbation: random fixed number of nodes and edges (same as $G_c$) are masked, denoted by \texttt{Randomnodeedgemask};
    \item Highest attention nodes perturbation: remove nodes with the highest attention weights(same number as $G_c$), hypothesizing they are structurally “most important”, denoted by \texttt{Higherattention}; 
\end{itemize}

\section{Initial Results} \label{sec:result}
Appendix.~\ref{sec:training} outlines the training configurations and results associated with the BioToolKG injection. Section~\ref{sec:explaincf} presents a preliminary analysis of the experimental results introduced in Section~\ref{experiment1}. Section~\ref{sec:baselineresult} compares our approach to baseline methods, as described in Section~\ref{baseline}.

\subsection{\texttt{CFFTLLMExplainer} Framework (A case study)} \label{sec:explaincf}

In this section, the initial graph $G$ is tailored as a transcript assembly pipeline as Figure.~\ref{fig:scallop}. Rather than generating the graph via LLM prompting, $G$ is explicitly defined based on standard workflows in real-world RNA-seq analysis. This design enables us to inject precise domain priors, ensure structural-level interpretability, and systematically analyze the influence of graph masking on LLM outputs. The use of precise pipeline structures and semantically coherent connections enhances the interpretability of the experiment. In contrast, pipelines generated by LLMs may have invalid or meaningless connections, thereby increasing the difficulty of subsequent explanation tasks. While future work may investigate generating such pipelines via prompting, our current focus is on counterfactual explanations based on a fixed structured input.

Following Equation~\ref{eq:overall}, the total loss $\mathcal{L}_{total}$ is instantiated with the following coefficient configuration: $\alpha = 400.0$ for semantic divergence, $\beta = 0.05$ for entropy regularization, $\gamma = 10.0$ for structure preservation, $\delta = 10.0$ for hard retention, and $\epsilon = 5.0$ for Laplacian smoothness. The structure sparsity term incorporates node and edge regularization weights $\lambda_V = 0.1$ and $\lambda_E = 0.5$, respectively, along with edge-type-specific coefficients: $\lambda_{\texttt{rels\_input}} = \lambda_{\texttt{rels\_output}} = 4.0$, $\lambda_{\texttt{rels\_download\_from}} = 0.5$, and $\lambda_{\texttt{END}} = 1.0$. To balance discrete thresholding and gradient stability, the Gumbel-softmax temperature is set to $\tau = 0.15$.

\begin{figure}[htbp]
    \centering
    \begin{subfigure}[t]{0.3\linewidth}
        \centering
        \includegraphics[width=\linewidth]{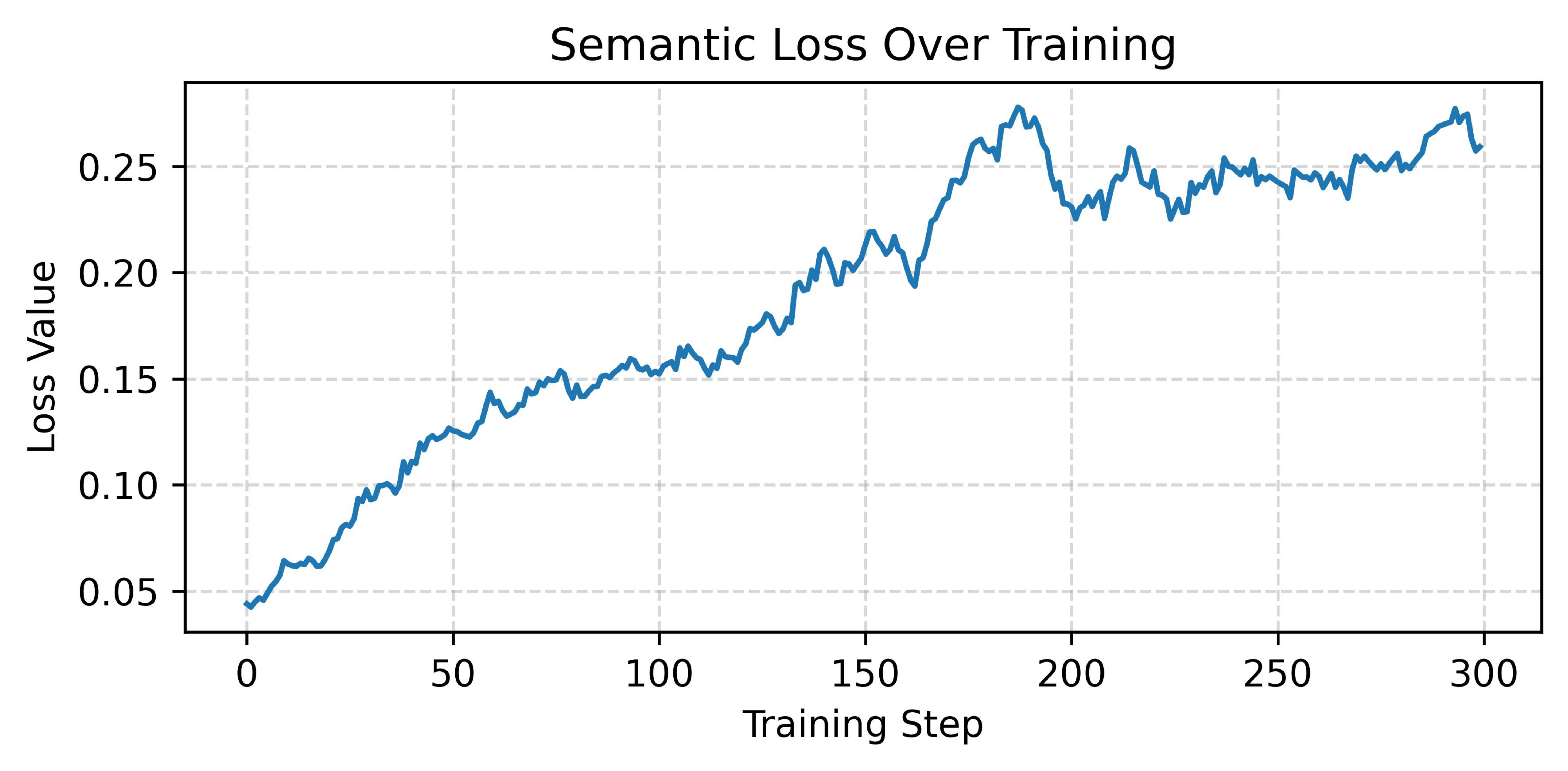}
        \caption{Semantic loss}
        \label{fig:semantic}
    \end{subfigure}
    \hfill
    \begin{subfigure}[t]{0.3\linewidth}
        \centering
        \includegraphics[width=\linewidth]{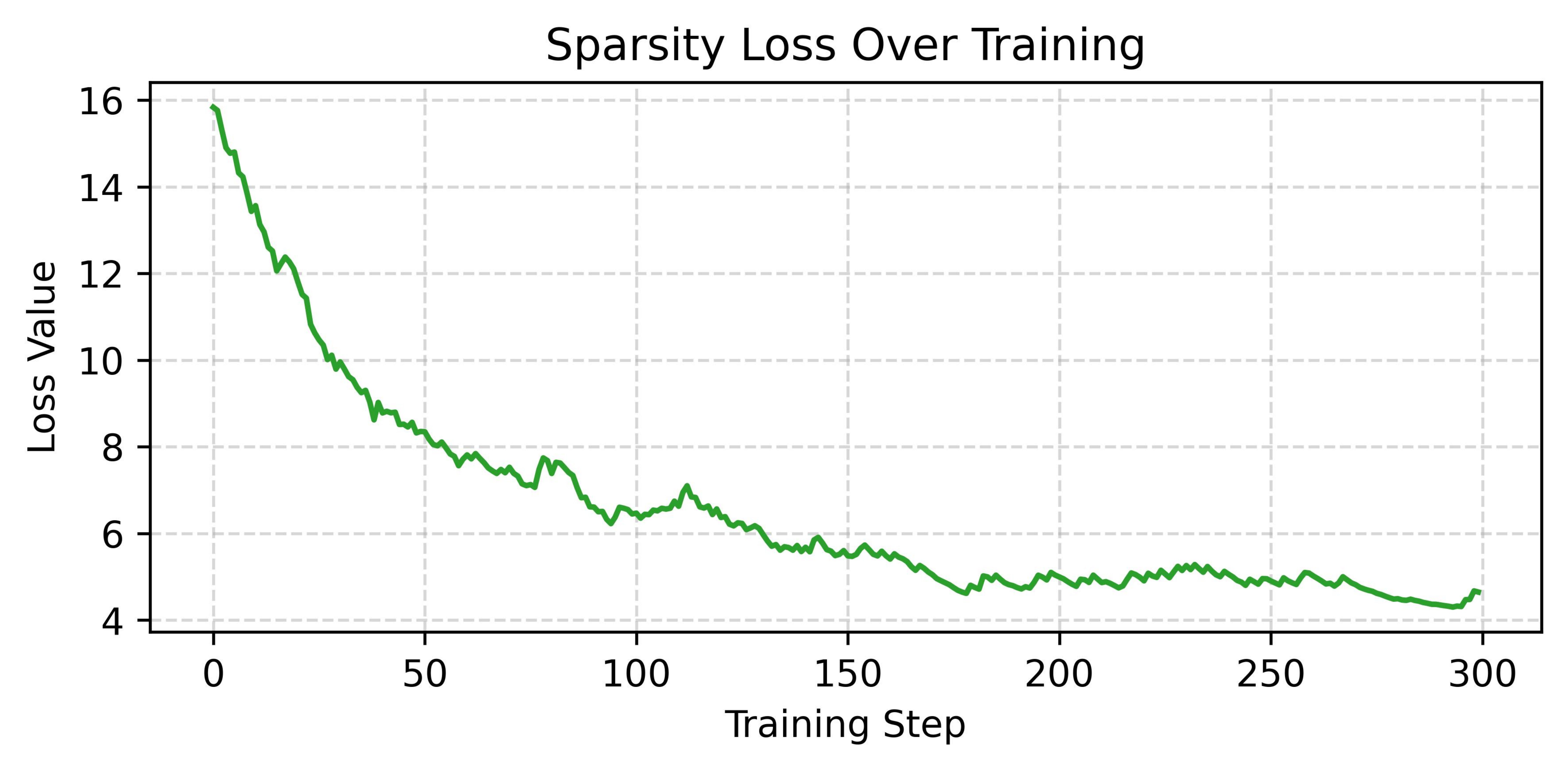}
        \caption{Sparsity loss}
        \label{fig:sparsity}
    \end{subfigure}
    \hfill
    \begin{subfigure}[t]{0.35\linewidth}
        \centering
        \includegraphics[width=\linewidth]{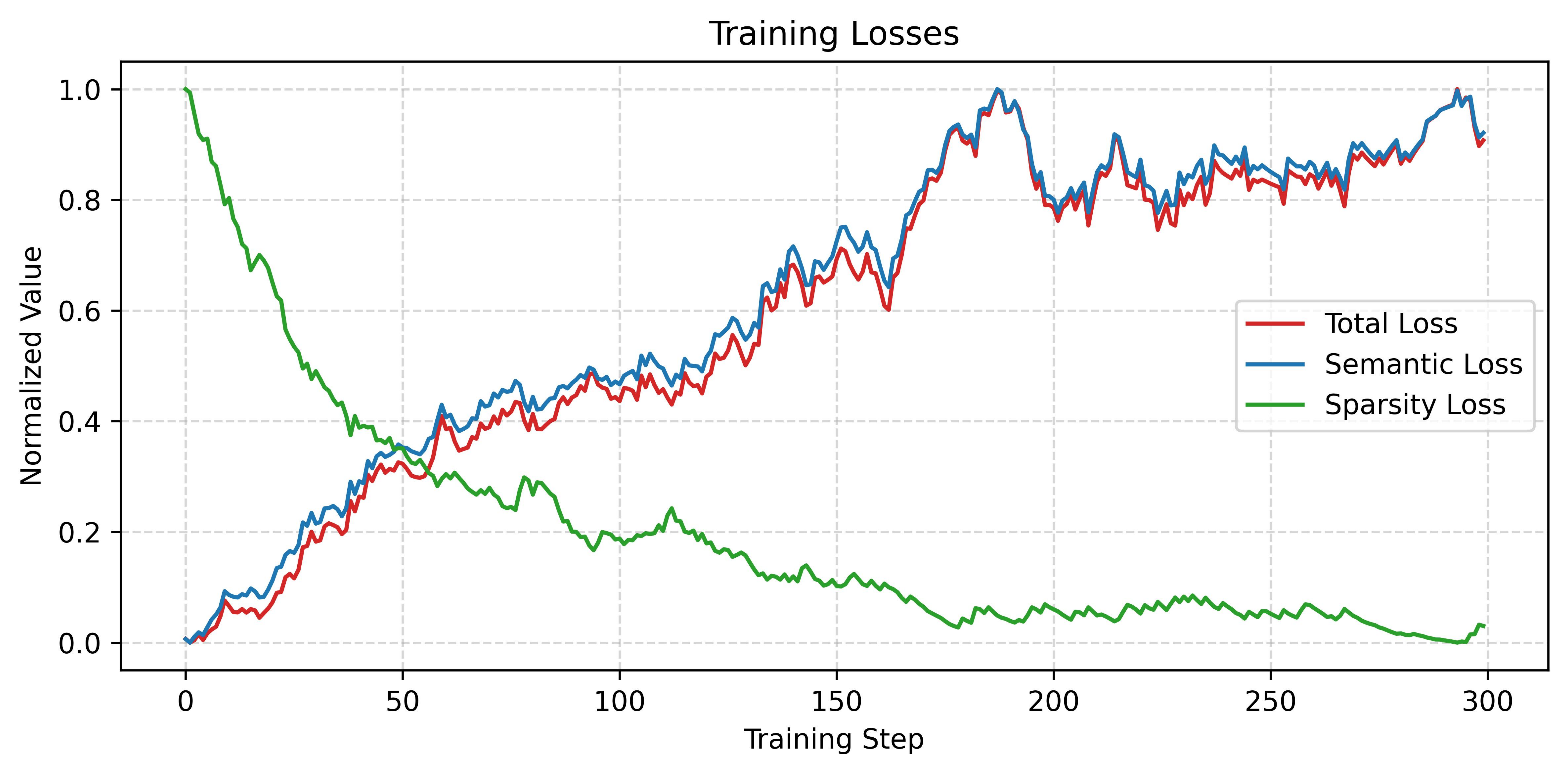}
        \caption{Overall loss after normalization}
        \label{fig:overall}
    \end{subfigure}

    \caption{Training loss curves including semantic loss, sparsity loss, and normalized overall loss.}
    \label{fig:all_losses}
\end{figure}

Figure~\ref{fig:all_losses} illustrates loss curves during soft masks training on the input graph $G$ shown in Figure~\ref{fig:scallop}. As shown in Figure~\ref{fig:semantic}, the semantic loss increases as structural perturbations are introduced, reflecting growing semantic divergence. Meanwhile, Figure~\ref{fig:sparsity} visualizes the sparsity loss, indicating that the graph becomes progressively sparser over time. In Figure~\ref{fig:overall}, we present the total loss with both semantic and sparsity components normalized, providing a comprehensive view of their combined influence. To better reflect the functional trends, all curves are smoothed using exponential moving average (EMA) with a decay factor of 0.9.

\begin{figure}[t]
    \centering
    \begin{subfigure}[t]{0.45\textwidth}
        \centering
        \includegraphics[width=\linewidth]{Figures_PDF/masked_graph.pdf}
        \caption{Masked Knowledge Graph}
        \label{fig:masked-graph}
    \end{subfigure}
    \hspace{0.5em}
    \begin{subfigure}[t]{0.22\textwidth}
        \centering
        \includegraphics[width=\linewidth]{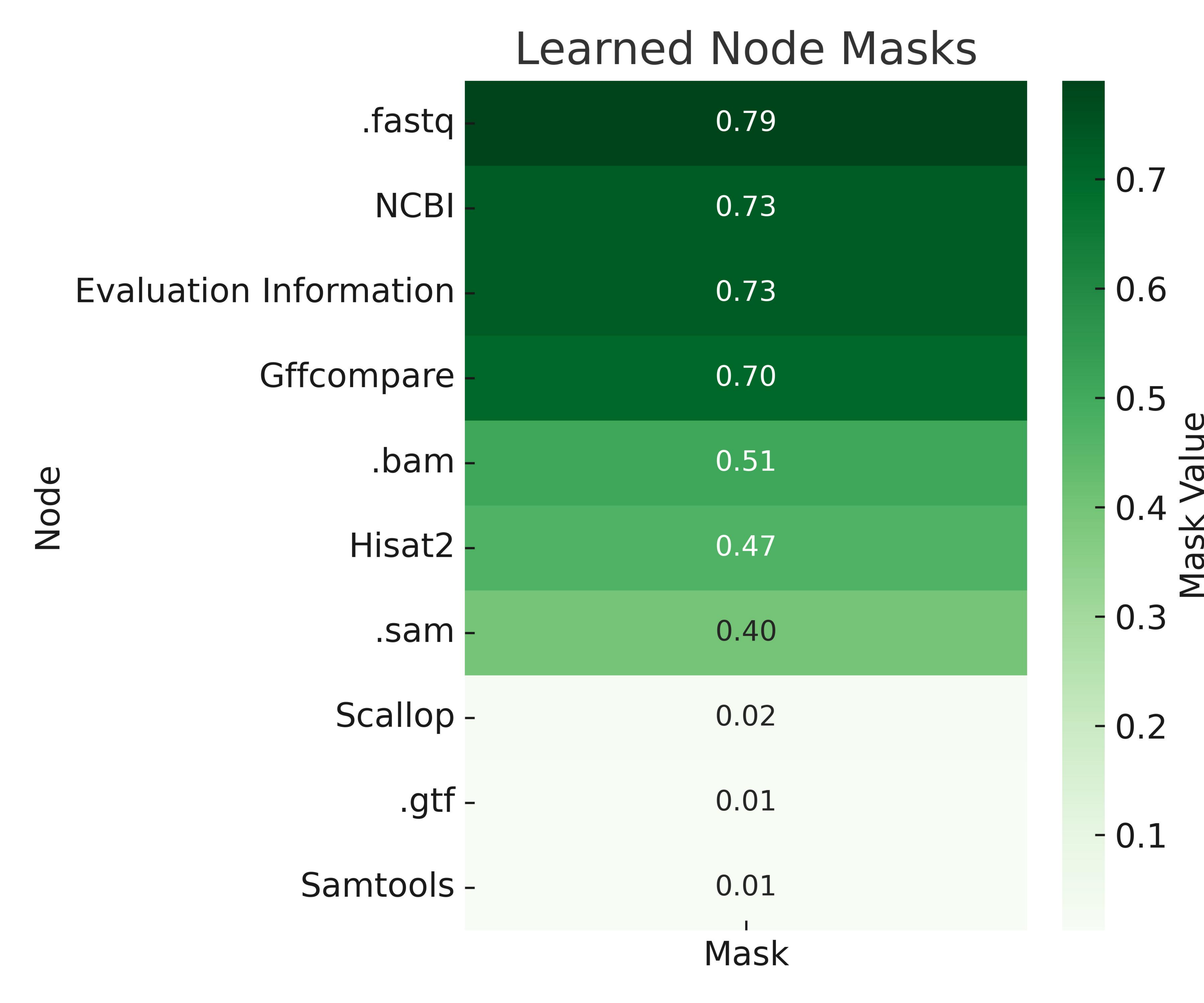}
        \caption{Nodes heatmap}
        \label{fig:node-mask}
    \end{subfigure}
    \hspace{0.5em}
    \begin{subfigure}[t]{0.22\textwidth}
        \centering
        \includegraphics[width=\linewidth]{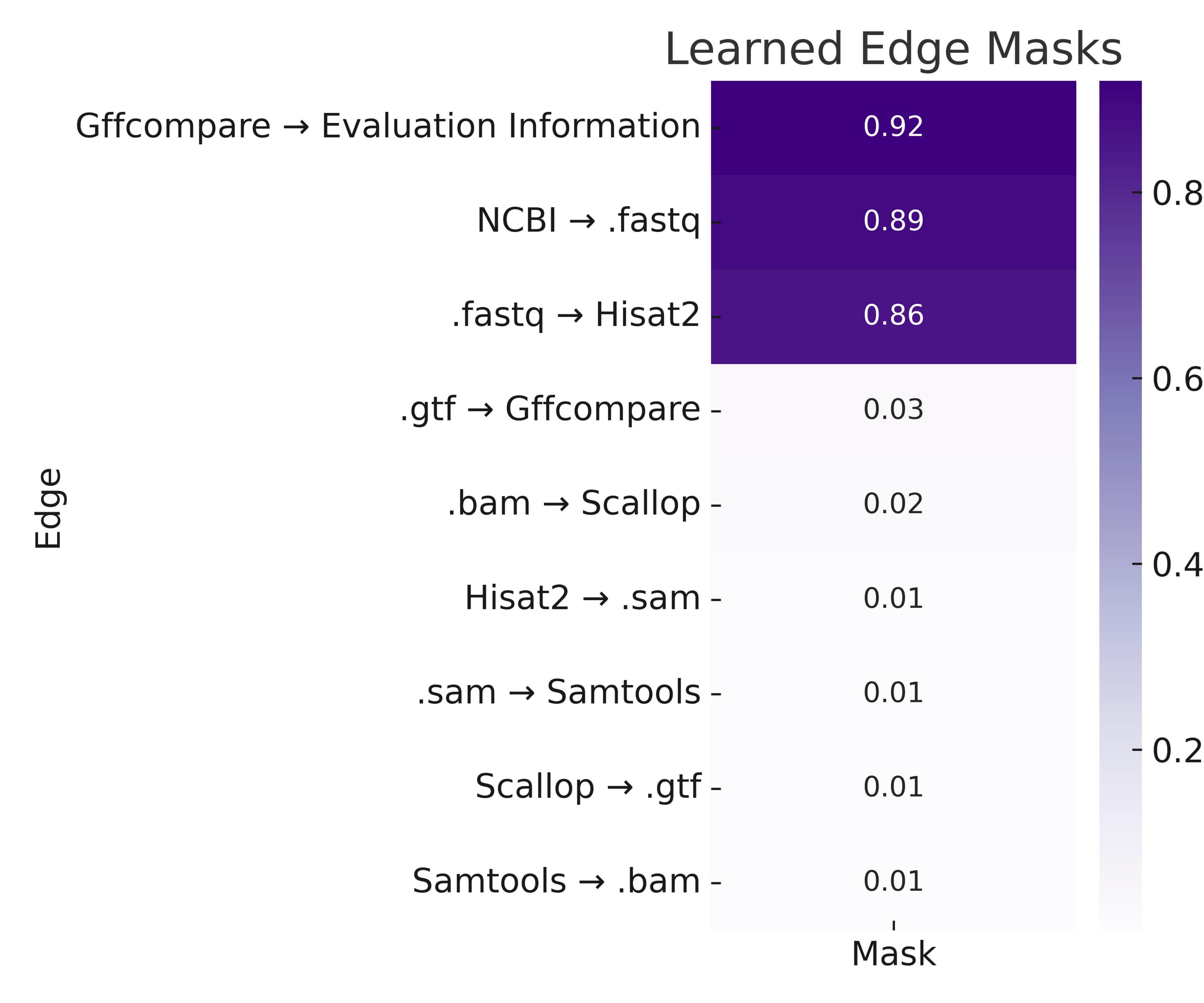}
        \caption{Edges heatmap}
        \label{fig:edge-mask}
    \end{subfigure}

    \caption{Visualization of the masked graph and corresponding node/edge mask heatmaps.}
    \label{fig:graph-masks}
\end{figure}

After the training process, \texttt{CFFTLLMExplainer} assigns a soft mask to each node and edge, representing its importance in the counterfactual subgraph $G_c$. It is important to note that Figure~\ref{fig:graph-masks} visualizes the retention weights of nodes and edges in $G_c$, darker colors indicate a higher likelihood of being preserved. However, this also implies that these elements are less critical in the original graph $G$, as their removal does not significantly affect the overall semantic structure, thus they are retained. Specifically, Figure~\ref{fig:masked-graph} illustrates the resulting graph structure after applying the learned masks, showing the retained nodes and edges in $G_c$. Figures~\ref{fig:node-mask} and~\ref{fig:edge-mask} further present the soft mask weights learned for each node and edge, respectively, indicating the probability of each being preserved in the counterfactual subgraph.

Following the experimental design outlined in Section~\ref{experiment1}, we obtain the outputs $f_{base}(G)$, $f_{ft}(G)$, and $f_{ft}(G_c)$ based on the original graph $G$ and the learned counterfactual subgraph $G_c$. In the subsequent analysis, we primarily focus on the bioinformatics tool entities mentioned in the outputs, as Tool-type nodes are assigned higher structural importance during mask training and carry greater interpretive weight in the pipeline.

\begin{figure}[htbp]
    \centering
    \begin{minipage}{\linewidth} 
        \centering
        \includegraphics[width=0.3\linewidth]{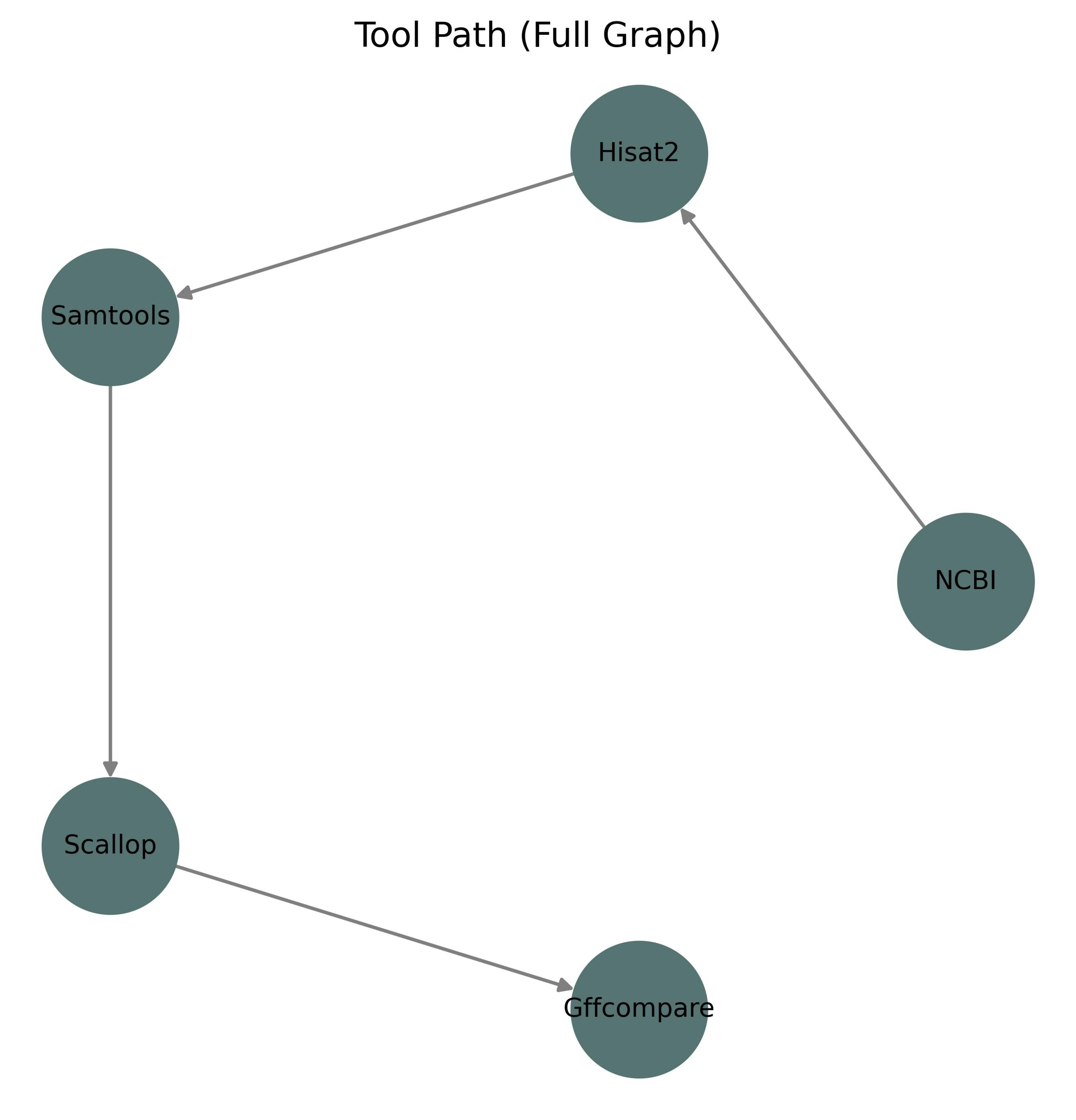}
        \includegraphics[width=0.3\linewidth]{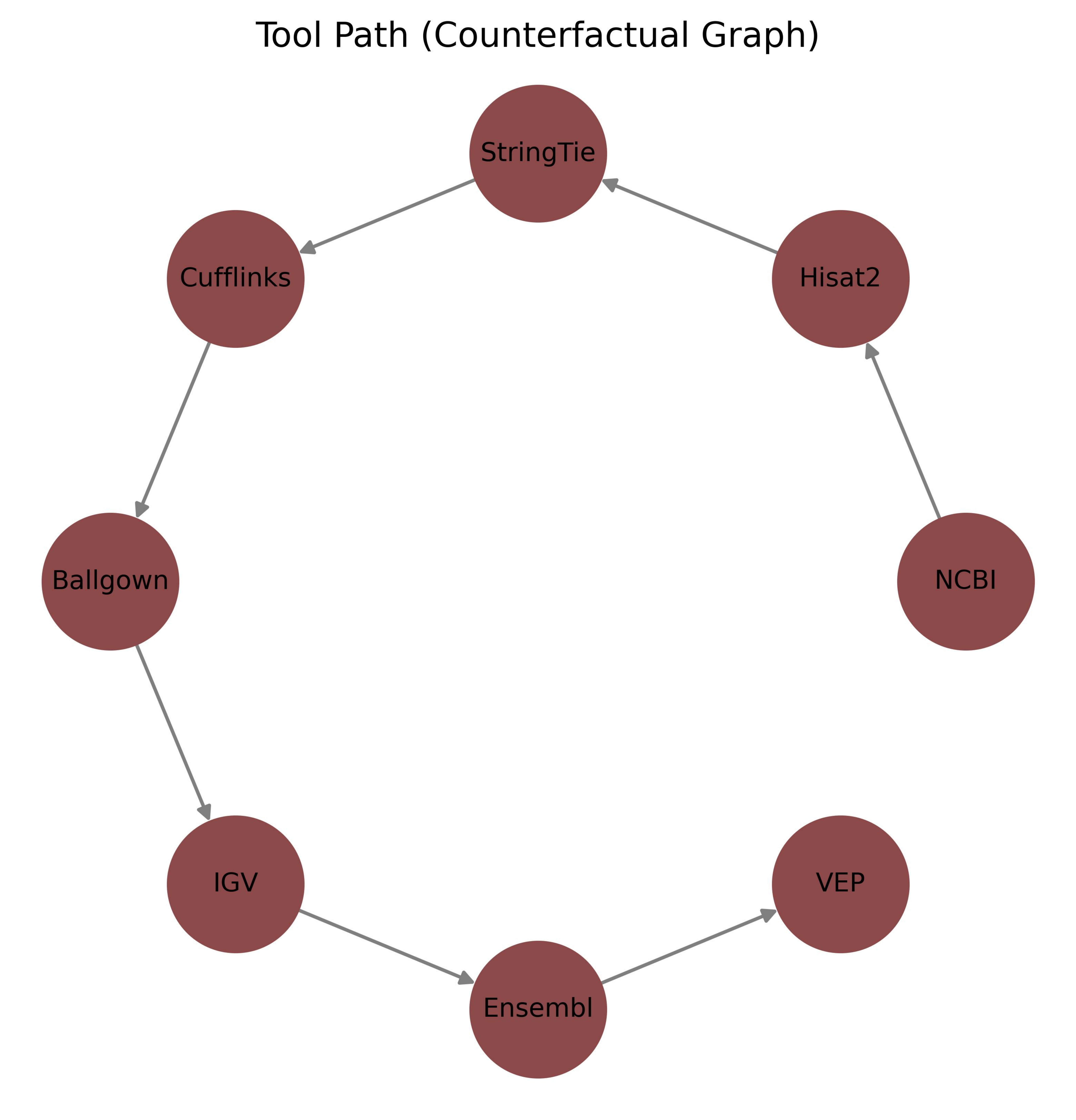}
    \end{minipage}
    \caption{Comparison: (left) full graph outcome via both baseline and fine-tuned model; (right) CF graph output via fine-tuned model.}
    \label{fig:graph-comparison}
\end{figure}

Figure~\ref{fig:graph-comparison} shows the toolchain extracted from the outcome of $f_{base}(G)$ and $f_{ft}(G)$. It is observed that both the baseline and fine-tuned models produce identical tool chain sequences when prompted with the full graph $G$, suggesting that the fine-tuning process preserves the semantic fidelity of the original pipeline. However, when prompted with the counterfactual graph $G_c$, the fine-tuned model generates a substantially different set of tools (see Figure~\ref{fig:graph-comparison}), demonstrating that the learned structural perturbations are sufficient to induce meaningful divergence in model behavior.

The structural and semantic behavior of the fine-tuned LLM is examined by comparing its outputs over the full graph $G$ and a counterfactual variant $G_c$. The analysis is conducted along three key dimensions (in Section~\ref{experiment1}):

\paragraph{Semantic Drift Analysis.}As shown in Figure~\ref{fig:graph-comparison}, the model produces significantly different toolchains in response to $G$ and $G_c$. Specifically, the tool set generated from $G$ includes \{\texttt{Hisat2}, \texttt{Samtools}, \texttt{Scallop}, and \texttt{Gffcompare}\}, while $G_c$ results in an entirely different set including \texttt{Ballgown}, \texttt{Cufflinks}, \texttt{StringTie}, \texttt{IGV}, \texttt{VEP}, and others. The computed Jaccard similarity between the two tool sets is only 0.1, indicating substantial deviation. This confirms that the structural perturbation induced by the learned masks leads to semantically distinct reasoning. The semantic distance between the full-graph output $f_{ft}(G)$ and the counterfactual output $f_{tf}(G_c)$ is measured by cosine dissimilarity, yielding a score of $0.544$. This suggests that the graph-level structural perturbation effectively triggers high-level semantic variation in model generation. Detailed metrics can be found at Table.~\ref{tab:semanticmetrics}.
\begin{table}[htbp]
    \caption{Semantic Drift Metrics}
    \label{tab:semanticmetrics}
    \centering
    \begin{tabular}{cccc}
        \toprule
        Jaccard & Edit Distance & Path Overlap & Cosine Similarity \\
        \midrule
        0.1018 & 6 & 0.25 & 0.5443 \\
        \bottomrule
    \end{tabular}
\end{table}

\paragraph{Attention Alignment Evaluation.}
Although the removed nodes, such as \texttt{Scallop} (0.0015), \texttt{Samtools} (0.0015), and \texttt{Gffcompare} (0.001), play pivotal roles in downstream reasoning, they consistently exhibit extremely low average attention weights (all below $0.002$). This discrepancy reveals a misalignment between the model's attention distribution and the true semantic importance of structural components. Such a mismatch aligns with previous observations that attention mechanisms, while integral to model architecture, may not reliably reflect the underlying decision-making process \cite{jain2019attention, wiegreffe2019attention}. These findings reinforce the need for structure-aware explanation approaches, as an overreliance on raw attention weights can obscure key elements in the input graph that are crucial for the model’s predictive behavior.

\begin{figure}[htbp]
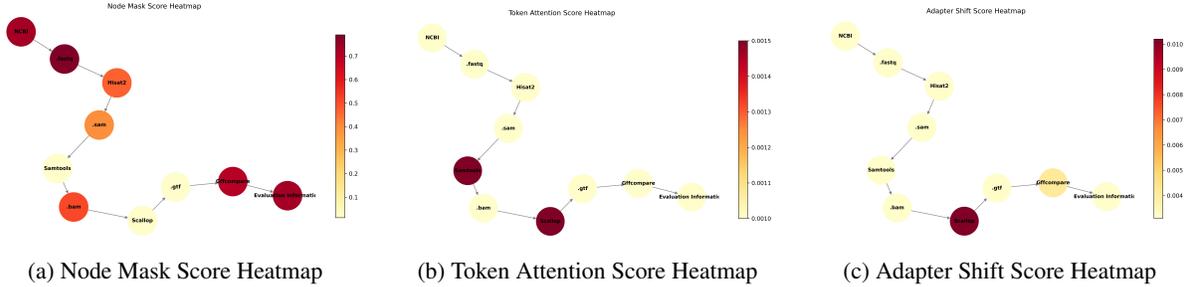

    \centering

    \begin{subfigure}[t]{0.3\textwidth}
        \centering
        \includegraphics[width=\linewidth]{Figures_PDF/node_mask.pdf}
        \caption{Node Mask Score Heatmap}
        \label{fig:maskedkg}
    \end{subfigure}
    \hspace{1em}
    \begin{subfigure}[t]{0.3\textwidth}
        \centering
        \includegraphics[width=\linewidth]{Figures_PDF/token_attention.pdf}
        \caption{Token Attention Score Heatmap}
        \label{fig:attention}
    \end{subfigure}
    \hspace{1em}
    \begin{subfigure}[t]{0.3\textwidth}
        \centering
        \includegraphics[width=\linewidth]{Figures_PDF/adapter_score.pdf}
        \caption{Adapter Shift Score Heatmap}
        \label{fig:adapter}
    \end{subfigure}

    \caption{Visual comparison of learned node masks, attention scores, and adapter shifts.}
    \label{fig:three heatmaps}
\end{figure}

\paragraph{Adapter Shift Probing.}Figure~\ref{fig:adapter} illustrates the adapter shift scores for selected tool-type tokens. Among them, \texttt{Scallop} exhibits the largest latent shift ($\|\Delta\| = 0.0102$), followed by \texttt{Gffcompare} ($\|\Delta\| = 0.0041$) and \texttt{Samtools} ($\|\Delta\| = 0.0031$). These results align with the training objective, as the fine-tuning data is centered around Scallop-related workflows. The magnitude of the shift thus provides an interpretable measure of token-level emphasis during LoRA adaptation. Furthermore, these scores can be compared with other interpretability signals, such as attention weights and node mask values—to triangulate the structural significance of each token. The discrepancy across signals reveals that some tokens with low attention or high retention likelihood in the counterfactual mask may still have strong adapter shifts, indicating a latent form of task-aware structural dependency.

Figure~\ref{fig:three heatmaps} presents the visualization results discussed earlier, illustrating three complementary interpretability signals on the transcript assembly graph. Figure~\ref{fig:maskedkg} shows the learned node mask scores from the \texttt{CFFTLLMExplainer}. Figure~\ref{fig:attention} displays the averaged attention weights over tool-type tokens in the input prompt. Figure~\ref{fig:adapter} depicts the adapter shift magnitudes derived from LoRA projection.

Overall, this experiment demonstrates that counterfactual structure masking offers a precise and faithful method for interpreting fine-tuned LLMs. It captures structural elements that most influence model behavior information that is often underrepresented in conventional attention visualizations.

\subsection{Baseline Comparison} \label{sec:baselineresult}
\begin{figure}
    \centering
    \includegraphics[width=0.6\columnwidth]{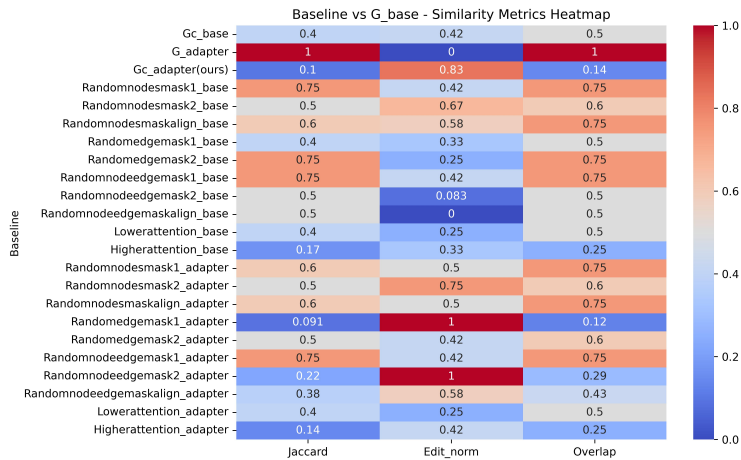}
    \caption{Baseline Experiments Heatmap}
    \label{fig:heatmap}
\end{figure}

Figure~\ref{fig:heatmap} shows the results in experiment 2 (Section~\ref{baseline}). Specifically, heatmap illustrating the similarity metrics between various baseline perturbation strategies and the original graph $G_{base}$. Each row corresponds to a baseline method applied to either the pretrained model ("\_\texttt{base}") or the LoRA fine-tuned model ("\_\texttt{adapter}"). It is important to note that in the \texttt{RandomNodeMask} and \\
\texttt{RandomNodeEdgeMask} experiments, the number of masked nodes and edges strictly follows that of the counterfactual toolchain $G_c$. However, the types of the masked nodes are not necessarily preserved. To address this, we introduce two additional baselines, \texttt{RandomNodeMaskAlign} and \texttt{RandomNodeEdgeMaskAlign}, which not only strictly match the number of masked nodes and edges in $G_c$, but also align with the types of the masked nodes (same in \texttt{Lowattention\_adapter} and \texttt{Higherattention\_adapter}). In parallel, the three evaluation metrics are designed to capture distinct dimensions of similarity: Jaccard quantifies semantic overlap between outputs, Edit Distance (normalized) measures structural deviations across different graph configurations, and Overlap assesses the consistency of the generated toolchains. Additional details regarding the baseline implementations can be found in Appendix~\ref{sec:appendixbaseline}.

From the figure, it is evident that \texttt{CFFTLLMExplainer} \\(\texttt{Gc\_adapter}) achieves a significantly low Jaccard similarity with respect to the original output, indicating a substantial semantic shift induced by the learned counterfactual structure. The relatively high \texttt{Edit\_norm} and low \texttt{Overlap} values further suggest that, under the structure of $G_c$, the model output undergoes not only semantic but also structural transformations. This supports the hypothesis that the fine-tuned LLM internalizes and adapts its structural knowledge during training. In contrast, most random mask baselines, while introducing considerable structural perturbations (as reflected by higher \texttt{Edit\_norm}), lead to only marginal changes in semantics and toolchain composition, as shown by higher Jaccard similarity and \texttt{Overlap} scores. This implies that such random modifications fail to establish a coherent mapping between structural changes and semantic behavior, and offer limited explanatory power.

Another meaningful and theoretically grounded baseline is \\ \texttt{Higherattention\_adapter}, which removes high-attention nodes of the same types as those in $G_c$. While it achieves moderately strong semantic shifts (e.g., Jaccard score of 0.14), its performance across all metrics remains consistently inferior to that of $G_c$. Moreover, the attention-based perturbation strategy suffers from limited interpretability, it is difficult to determine whether the observed output variations stem from the removal of high-attention nodes themselves or from the disruption of their associated structural context.

Moreover, in many cases, the resulting graphs from both random and attention-based perturbations exhibit disconnected or biologically implausible structures. These graphs deviate from domain-specific bioinformatics workflows, making it difficult to extract coherent toolchain semantics. In contrast, $G_c$ maintains biological plausibility while effectively inducing targeted semantic divergence, showcasing its advantage in producing interpretable and structurally grounded counterfactual explanations.

\section{Future Works} \label{sec:future}
While the case study have demonstrated that counterfactual graph masking can effectively reveal structure-sensitive behavior in the model, several directions remain for future exploration.

A promising future direction is to extend the current graph-based analysis to a prompt-driven setting by constructing instruction-style prompts from BioToolKG subgraphs. This would enable evaluation of semantic drift under structural perturbations at the prompt level. Future work will focus on formalizing this framework and designing improved semantic similarity metrics. Additionally, integrate human-feedback constraints are planned to use into the \texttt{CFFTLLMExplainer}, such as domain-specific path constraints or tool dependencies. Finally, while this study focuses on the biomedical domain, our framework is broadly applicable to other structured knowledge environments such as legal reasoning or scientific workflow modeling.

\section{Conclusion} \label{sec:conclusion}
We propose \texttt{CFFTLLMExplainer}, a counterfactual-based interpretability framework to analyze how fine-tuned LLMs process structured knowledge graph inputs. By introducing BioToolKG and a differentiable graph masking approach, we demonstrate that LoRA adapters encode structural biases that can be revealed through semantic drift induced by minimal graph perturbations.

Through visual and quantitative analyses, we identify structural components most influential to model outputs and show that conventional attention scores do not always capture these dependencies. Our findings offer new insights into the internal behavior of fine-tuned LLMs, emphasizing the importance of structure-aware explanation methods.

This work provides a foundation for future investigations into prompt-level semantic sensitivity, user-guided counterfactual control, and cross-domain generalizability of structural interpretability in LLMs.\textbf{}

\bibliographystyle{unsrt}  
\bibliography{references}

\appendix

\section{Knowledge Graph Injection} \label{sec:training}
We fine-tune the DeepSeek-R1-Distill-Llama-8B model using the QLoRA framework with 4-bit quantization and low-rank adapter injection. Instruction-style training data are tokenized and formatted into causal language modeling prompts. A LoRA adapter is trained over three epochs using standard Transformer optimization settings, and the resulting adapter weights are exported for subsequent inference and interpretability analysis. Table~\ref{tab:config} shows more detailed parameters and configurations.

\begin{table}[htbp]
\centering
\footnotesize
\begin{tabular}{>{\centering\arraybackslash}p{4.0cm} >{\centering\arraybackslash}p{3.8cm}}  
\toprule
\textbf{Category} & \textbf{Setting} \\
\midrule
\multicolumn{2}{c}{\textbf{LoRA Configuration}} \\
\midrule
LoRA rank ($r$) & 8 \\
LoRA scaling factor ($\alpha$) & 32 \\
LoRA dropout rate & 0.05 \\
Applied to modules & \texttt{q\_proj}, \texttt{k\_proj}, \texttt{v\_proj}, \texttt{o\_proj} \\
Bias adaptation & None \\
Task type & Causal Language Modeling \\
\midrule
\multicolumn{2}{c}{\textbf{Quantization Settings (QLoRA)}} \\
\midrule
Quantization precision & 4-bit \\
Quantization type & NF4 (NormalFloat 4-bit) \\
Computation data type & \texttt{float16} \\
Double quantization & Enabled \\
\midrule
\multicolumn{2}{c}{\textbf{Training Hyperparameters}} \\
\midrule
Per-device batch size & 4 \\
Gradient accumulation steps & 2 \\
Effective batch size & 8 \\
Number of training epochs & 3 \\
Learning rate & 2e-4 \\
Mixed precision & \texttt{float16} (training), \texttt{bfloat16} \\
Flash Attention & Disabled \\
\bottomrule
\end{tabular}
\caption{Configuration and training settings of QLoRA fine-tuning.}
\label{tab:config}
\end{table}


\begin{figure}[htbp]
    \centering
    \includegraphics[width=0.6\columnwidth]{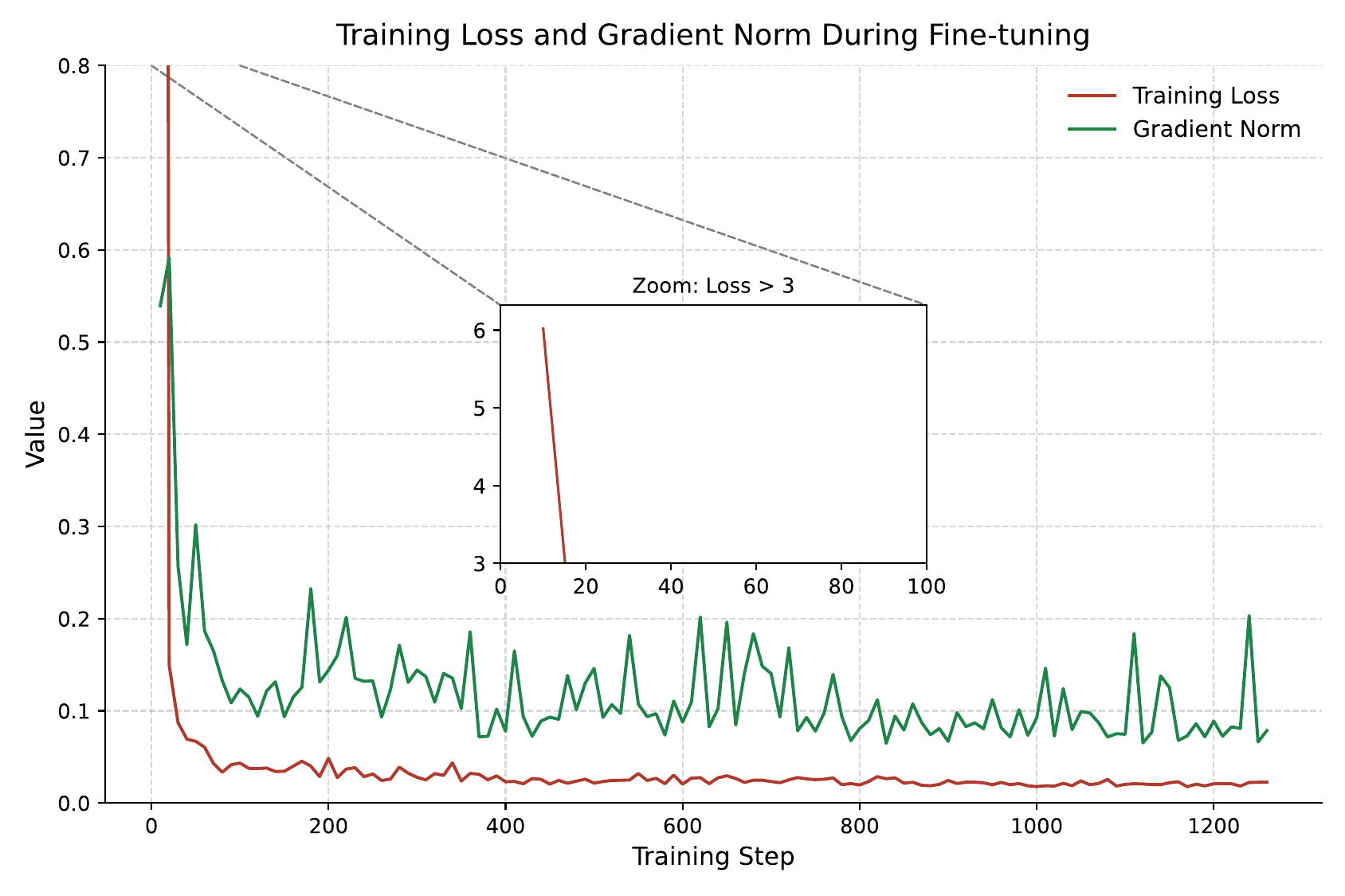}
    \caption{Loss and gradient norm curve for LoRA training}
    \label{fig:trainingcurve}
\end{figure}

The adapter introduces only 6.8M trainable parameters out of 8.0B total, yielding a trainable ratio of merely $0.0848\%$. As shown in Figure~\ref{fig:trainingcurve}, the training loss decreases sharply in the early stages and converges to a stable level below 0.05 within a few hundred steps, indicating successful knowledge injection into the adapter. The gradient norm exhibits natural oscillation throughout training. Such fluctuations are typical in parameter-efficient setups, where a small number of parameters carry most of the gradient updates.

\section{Baseline experiment details} \label{sec:appendixbaseline}

\begin{figure}[htbp]
    \centering
    
    \includegraphics[width=0.2\linewidth]{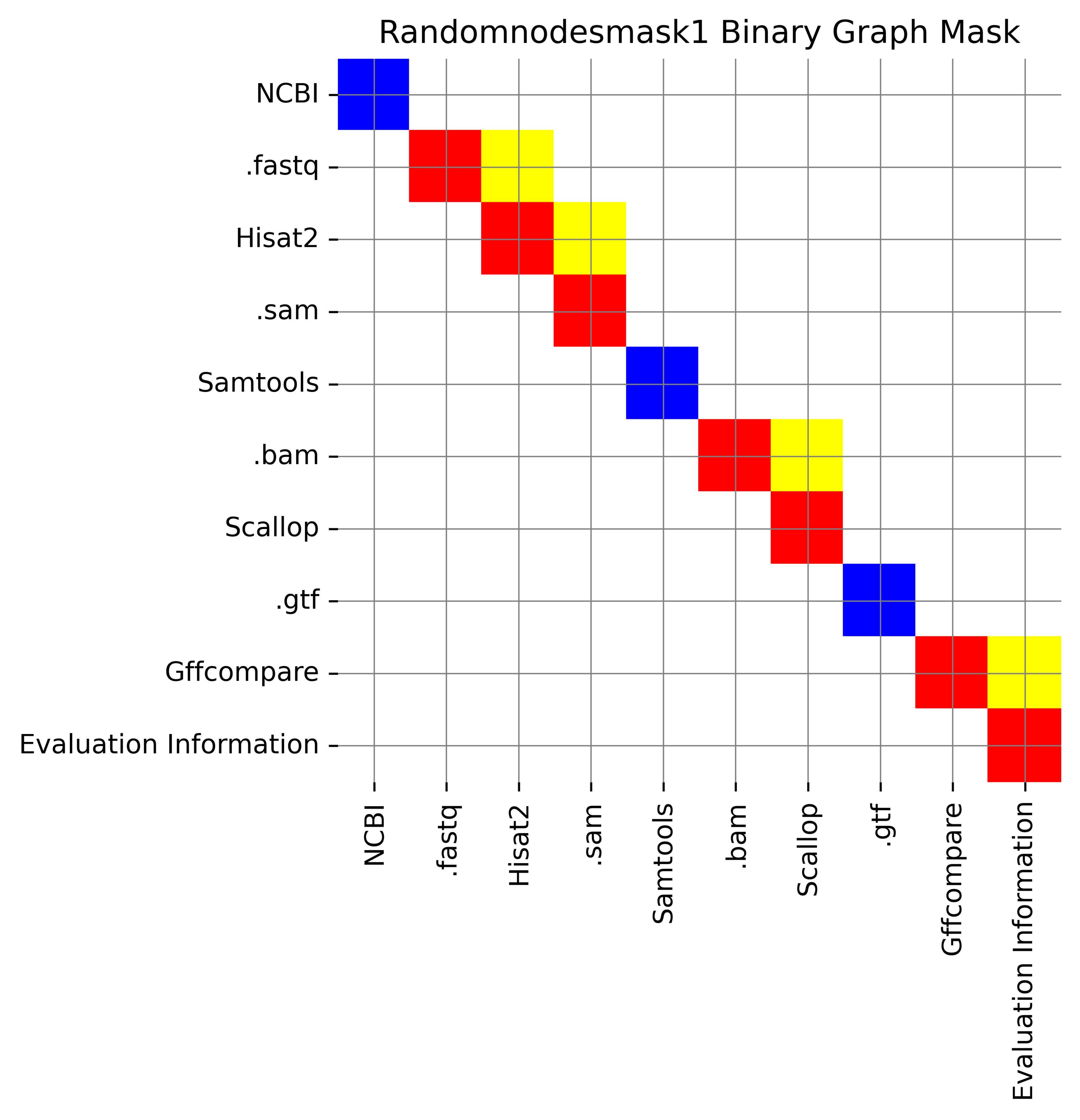}%
    \includegraphics[width=0.2\linewidth]{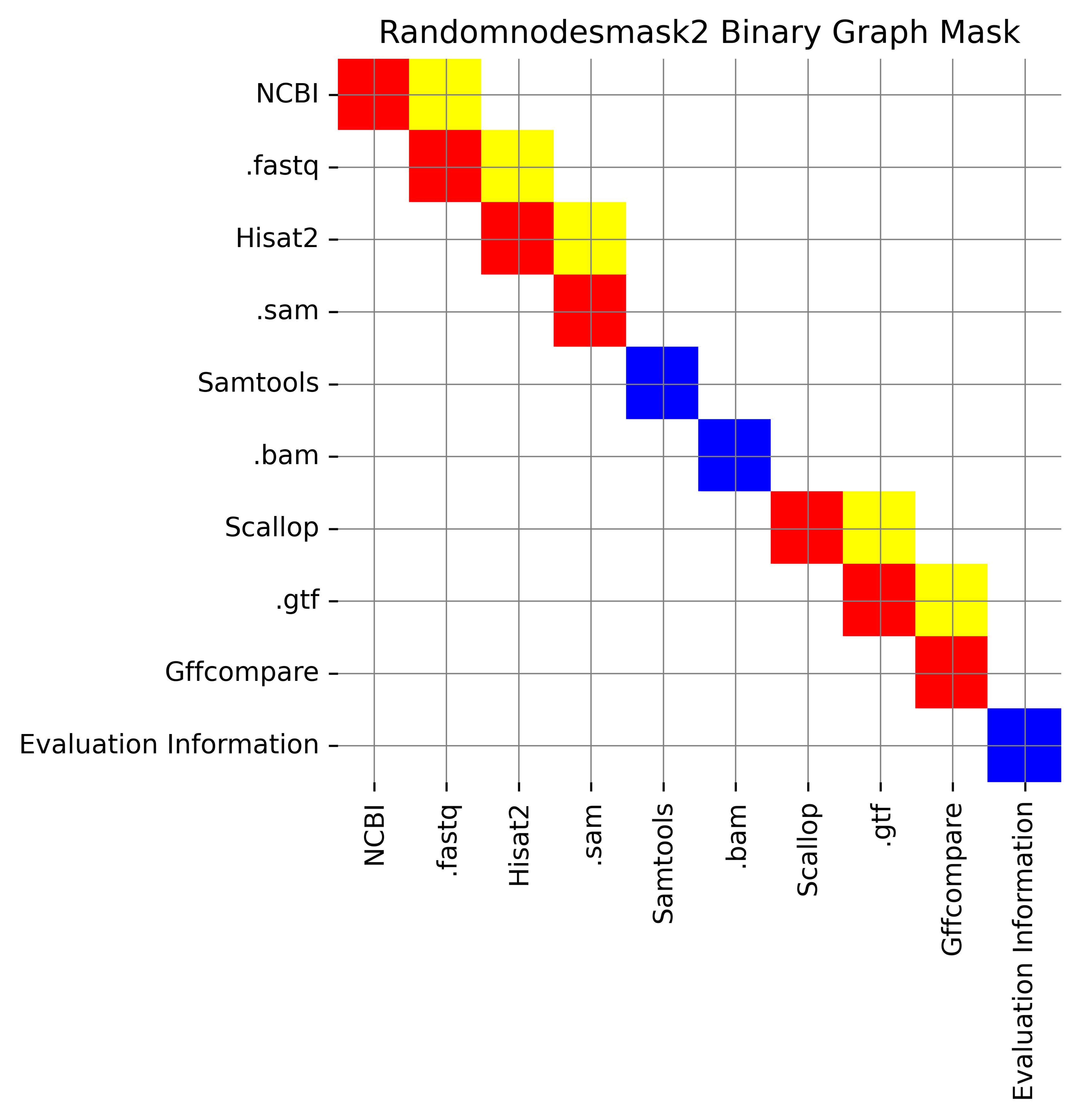}%
    \includegraphics[width=0.2\linewidth]{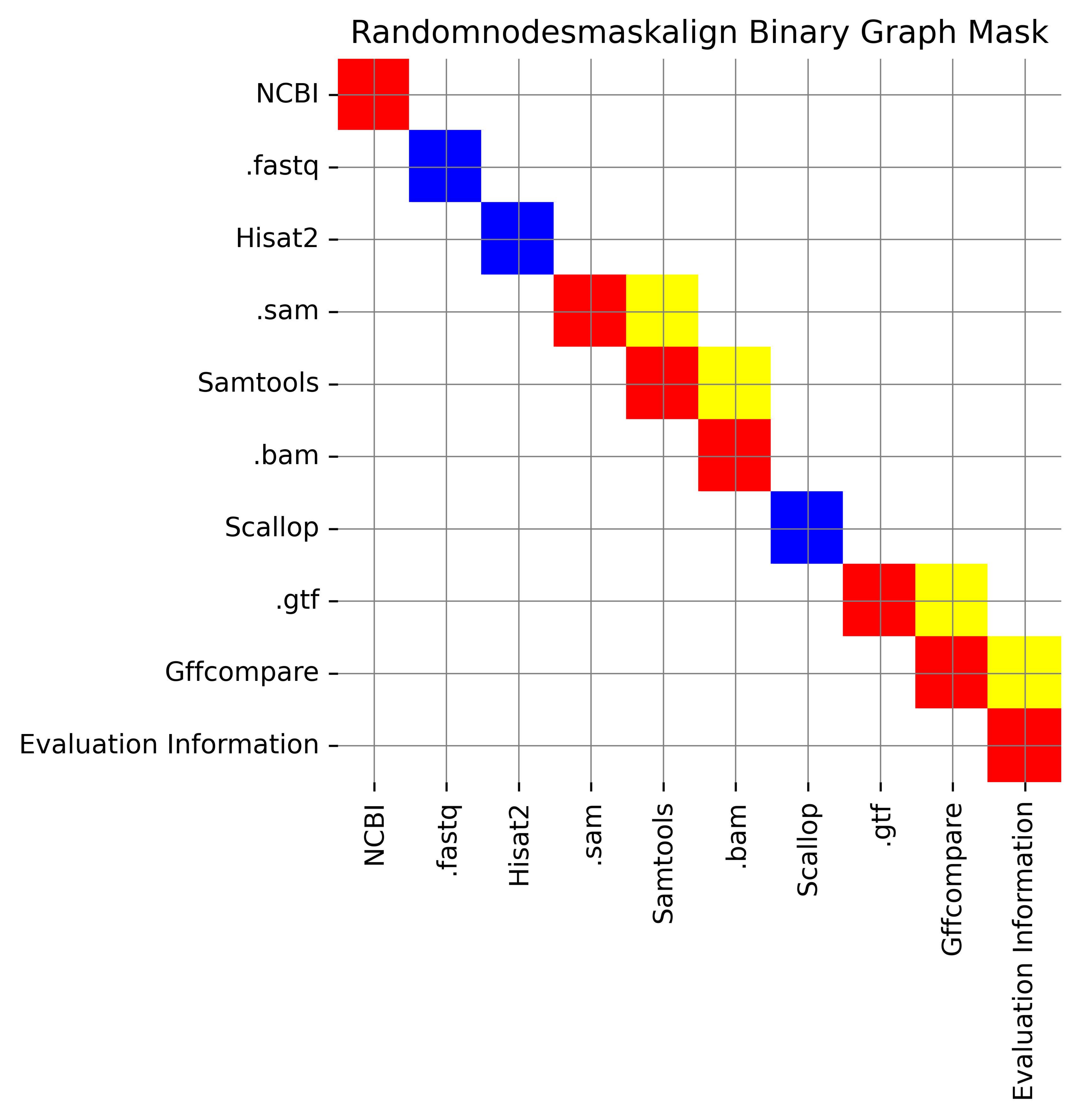}

    \vspace{0.5em}

    \includegraphics[width=0.2\linewidth]{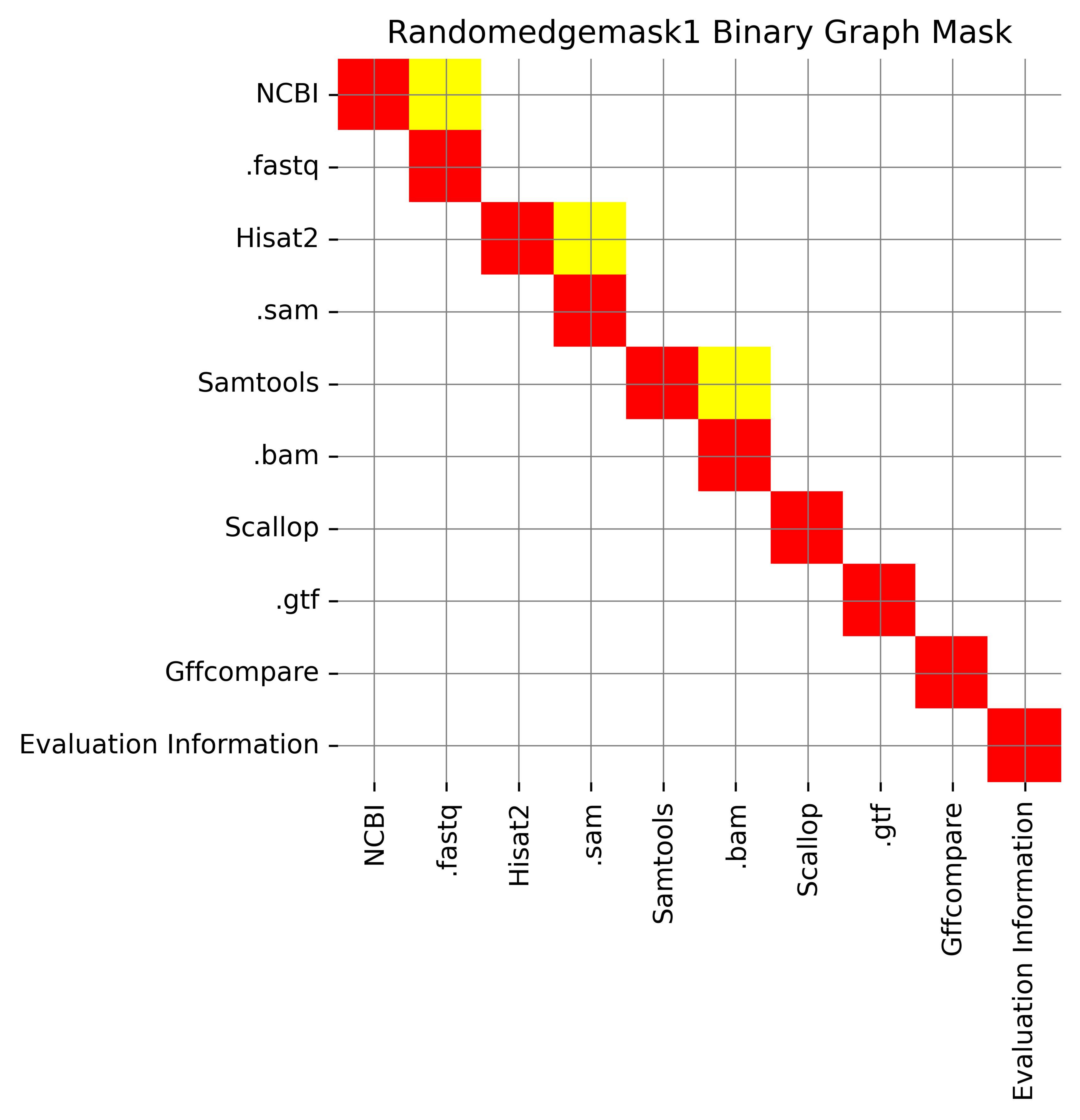}%
    \includegraphics[width=0.2\linewidth]{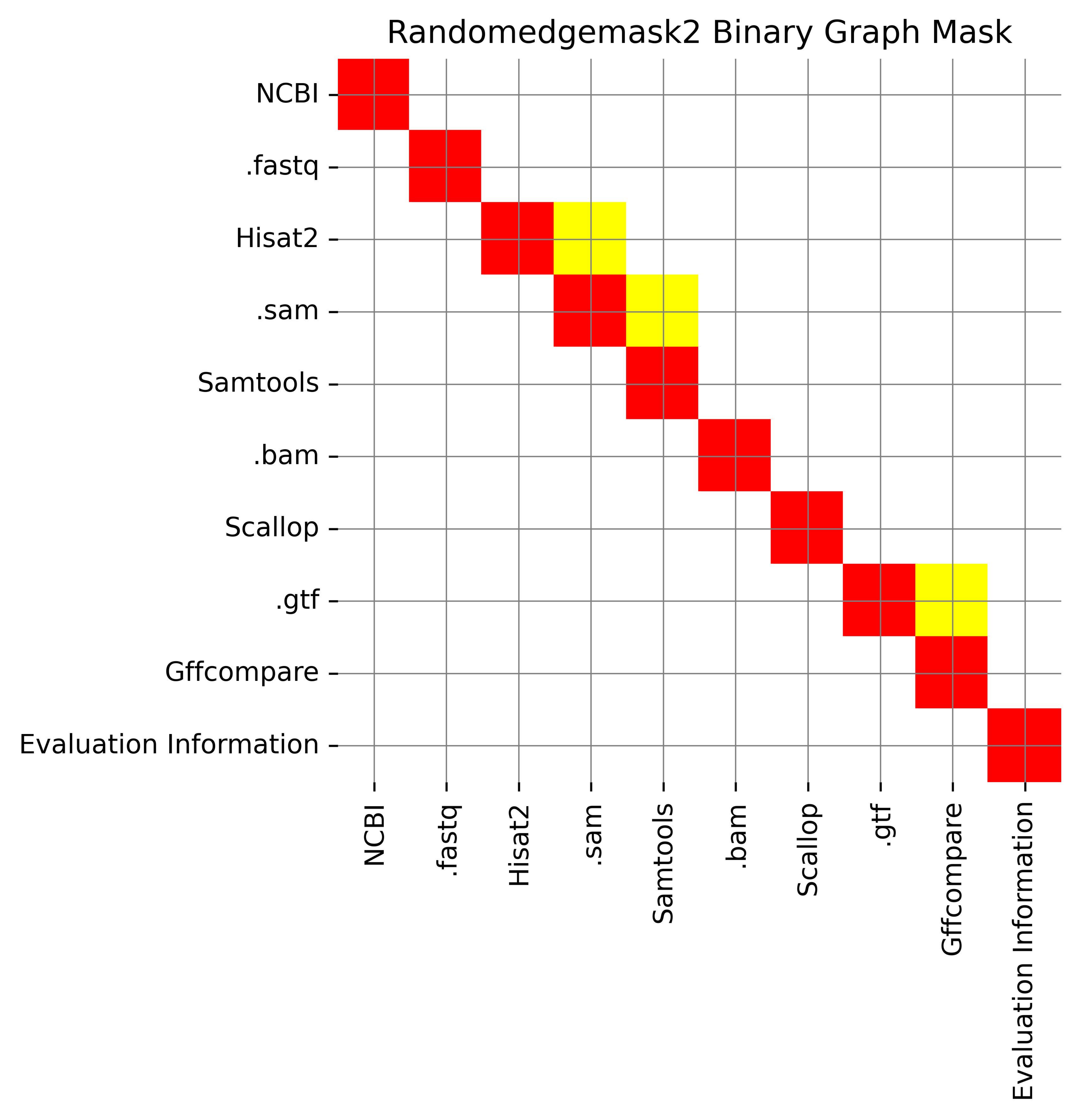}

    \vspace{0.5em}

    \includegraphics[width=0.2\linewidth]{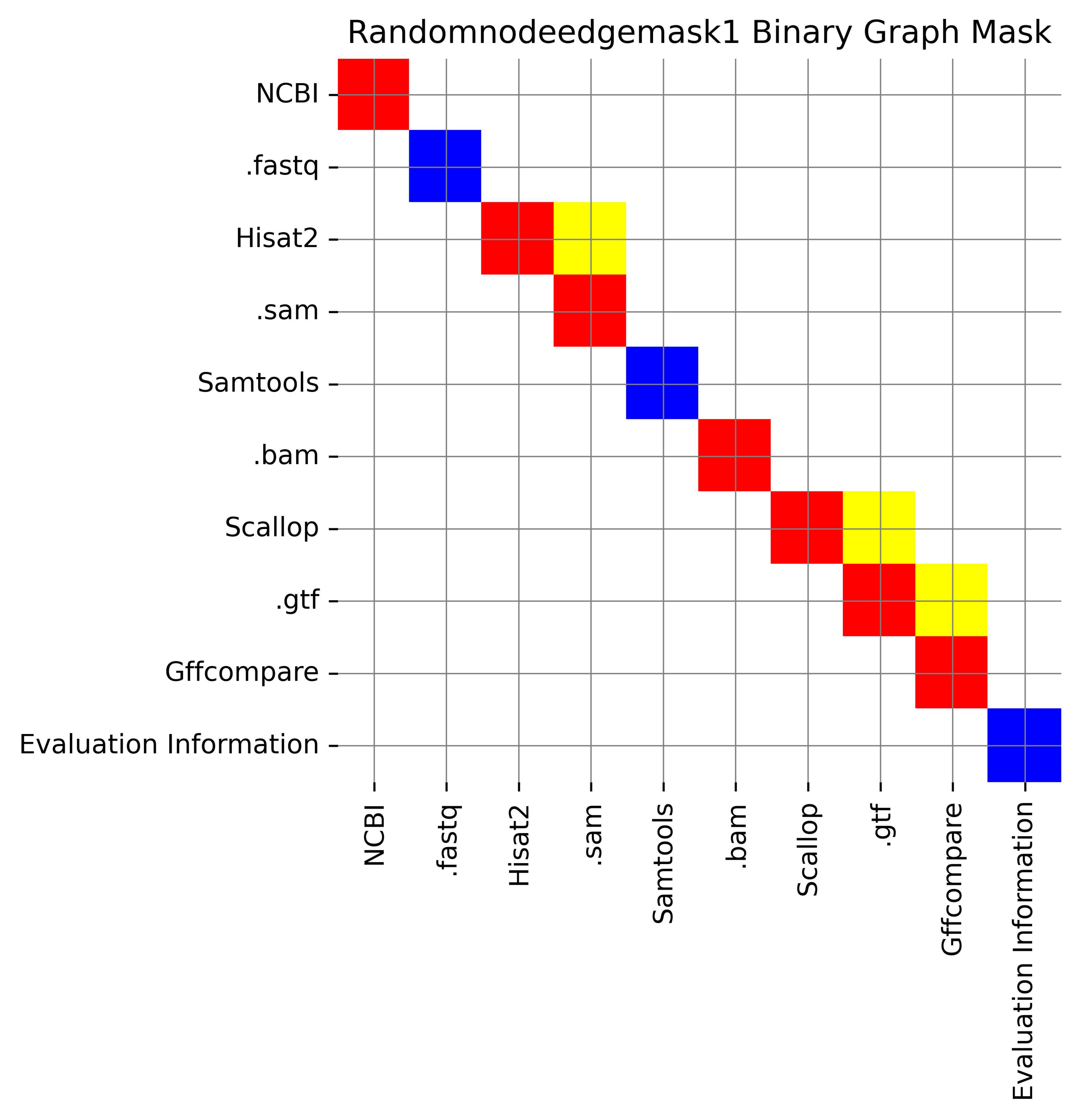}%
    \includegraphics[width=0.2\linewidth]{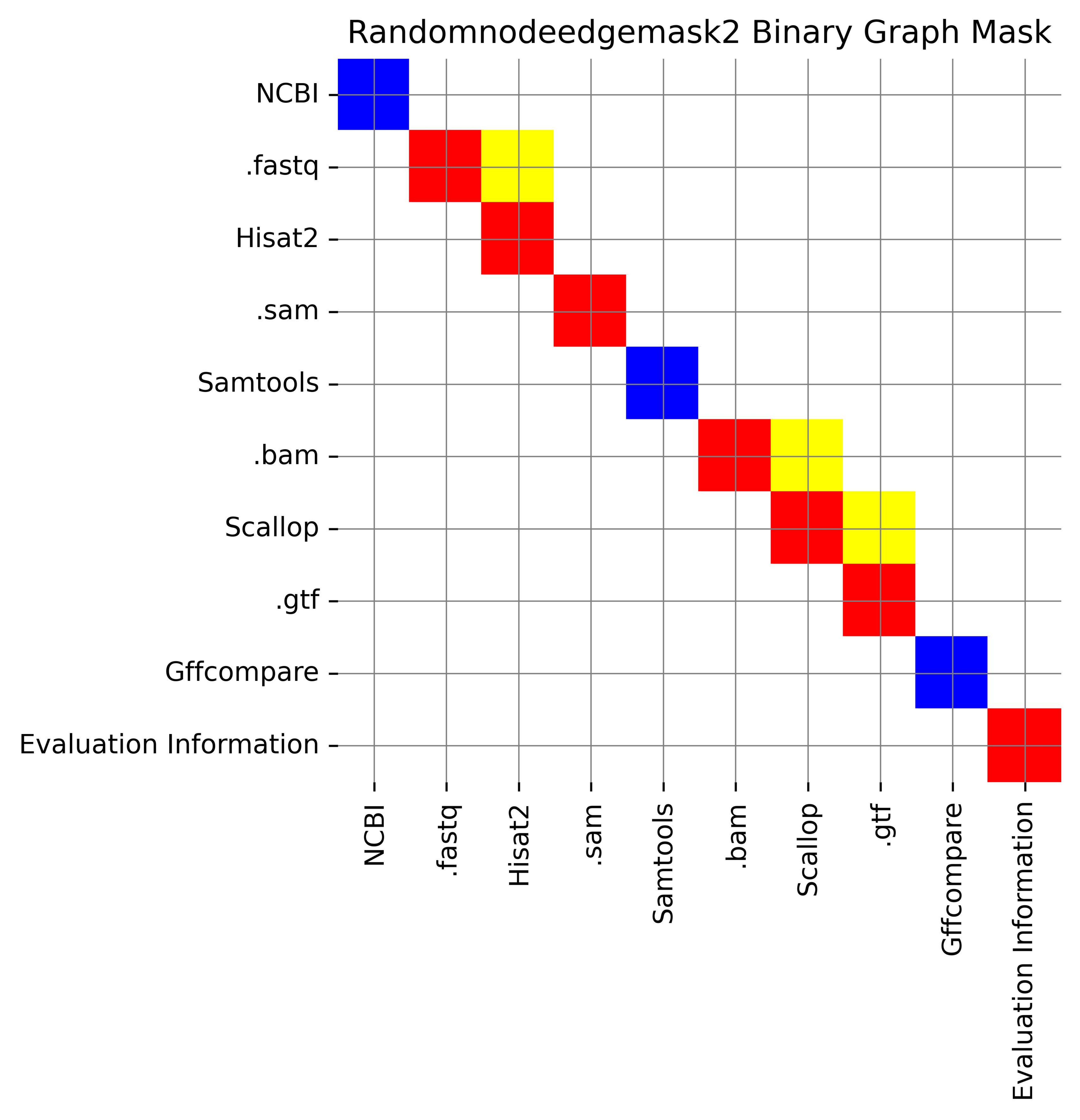}%
    \includegraphics[width=0.2\linewidth]{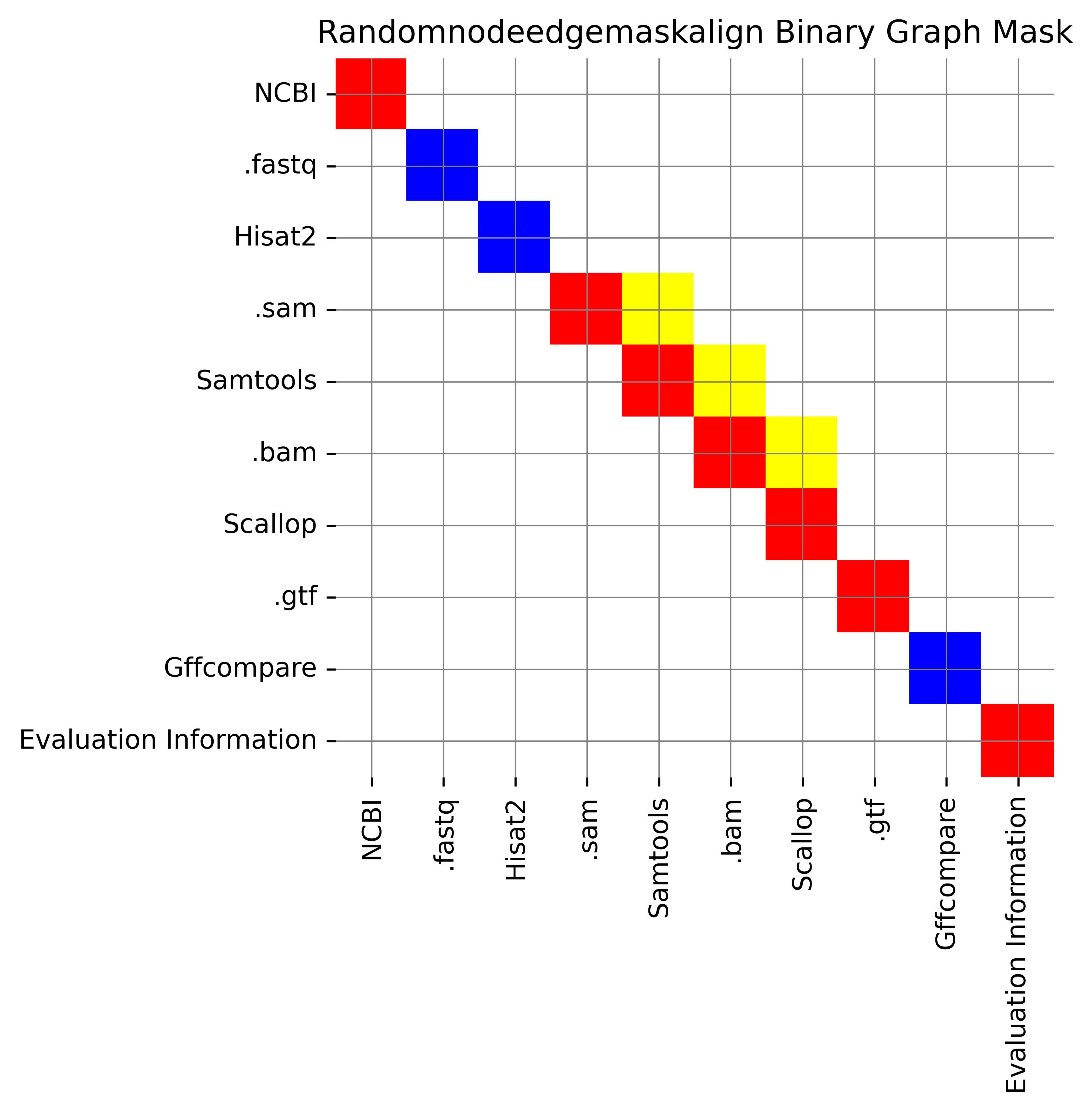}

    \vspace{0.5em}

    \includegraphics[width=0.2\linewidth]{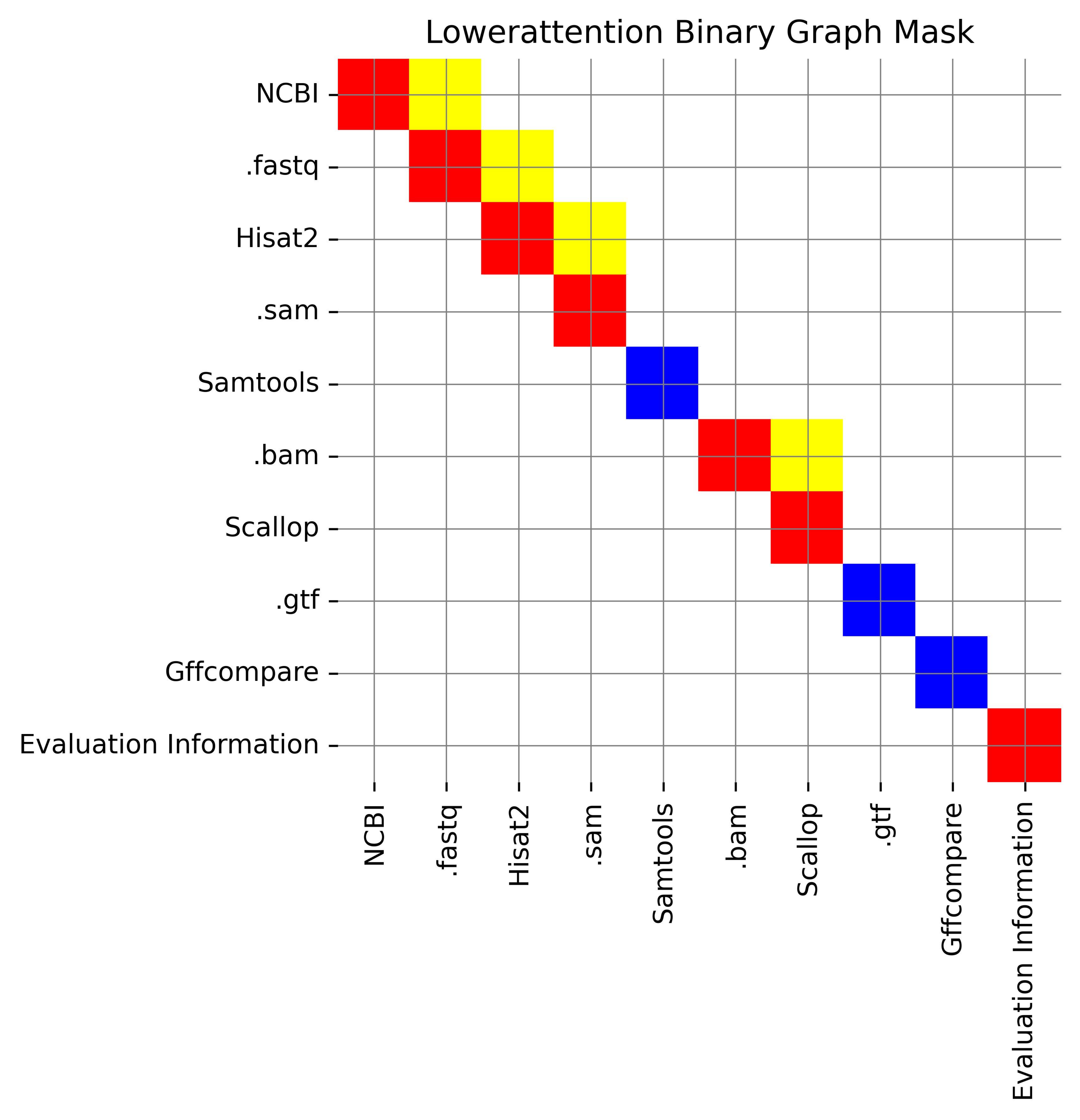}%
    \includegraphics[width=0.2\linewidth]{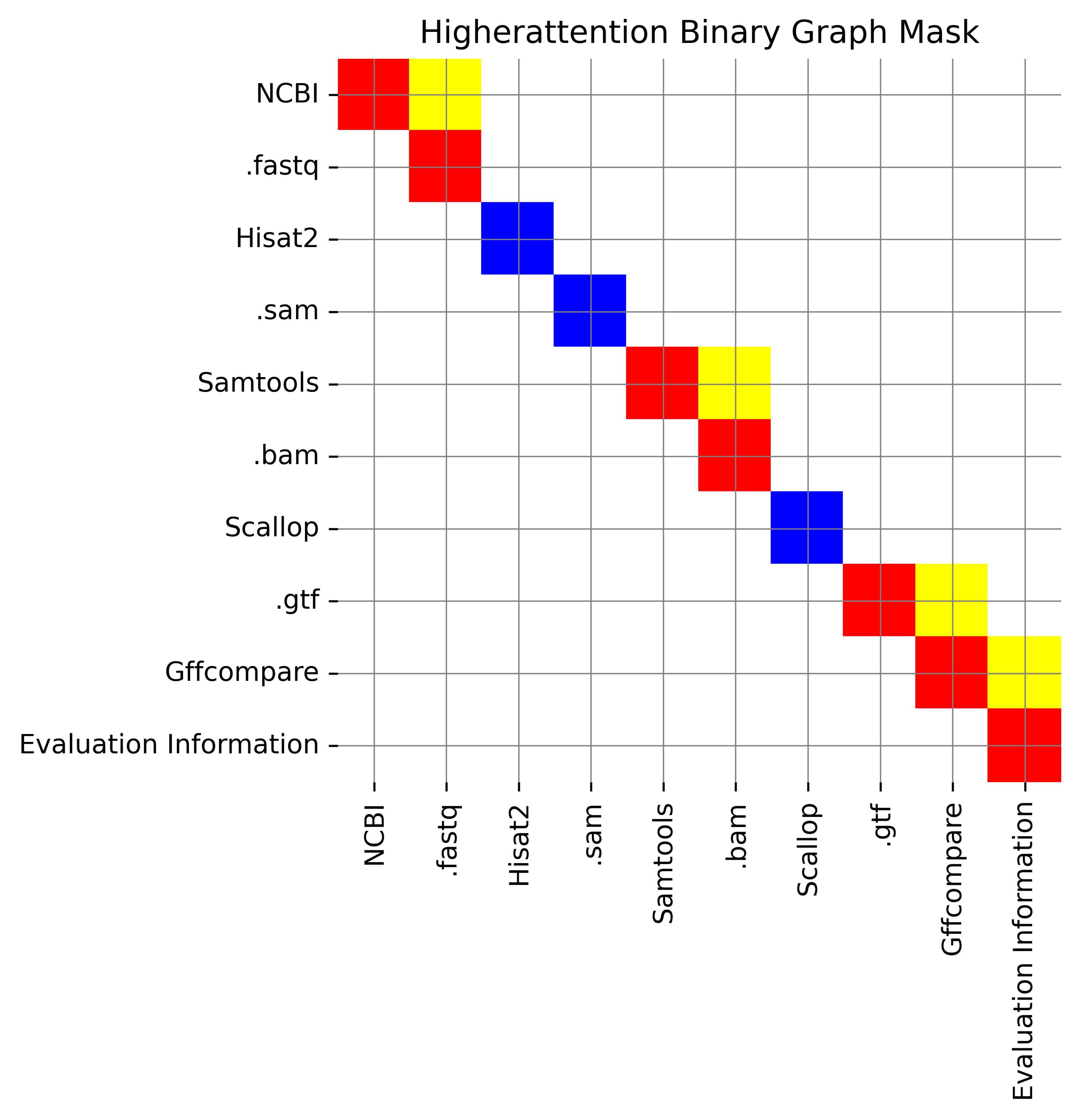}

    \caption{Binary Graph Masks under Different Perturbation Strategies}
    \label{fig:all_masks}
\end{figure}

Figure~\ref{fig:all_masks} presents adjacency matrix heatmaps illustrating the retention and removal of nodes and edges in each perturbed graph in Section~\ref{baseline}. The colors on the diagonal are used to indicate the retention status of each node. \textcolor{red}{Red} denotes retention, \textcolor{blue}{Blue} represents removal. The colors used for edges indicate their retention status. \textcolor{yellow}{Yellow} represents retained edges, while \textcolor{white}{White} indicates deleted edges.

Since the graph is directed, the adjacency matrix is asymmetric, meaning that the relationship between node $i$ and node $j$ may not be identical to the relationship between node $j$ and node $i$. Therefore, the values of edge retention may differ depending on the direction of the edge.

\begin{table}[htbp]
\centering
\scriptsize
\begin{tabular}{>{\arraybackslash}p{3.5cm} >{\arraybackslash}p{4.5cm}}  
\toprule
\textbf{Experiment Name} & \textbf{Extracted Tool Chain} \\
\midrule
G\_base & Gffcompare, Hisat2, Samtools, Scallop \\
Gc\_base & NCBI, Hisat2, Gffcompare \\
G\_adapter & Gffcompare, Hisat2, Samtools, Scallop \\
\rowcolor{gray!20} \textbf{Gc\_adapter (Ours)} & \textbf{Ballgown, Cufflinks, Ensembl, Hisat2, IGV, StringTie, VEP} \\
Randomnodesmask1\_base & Hisat2, Scallop, Gffcompare \\
Randomnodesmask2\_base & NCBI, Hisat2, Scallop, Gffcompare, StringTie \\
Randomnodesmaskalign\_base & NCBI, Hisat2, Scallop, Gffcompare \\
Randomedgemask1\_base & NCBI, Hisat2, Samtools \\
Randomedgemask2\_base & Hisat2, Samtools, Gffcompare \\
Randomnodeedgemask1\_base & Hisat2, Scallop, Gffcompare \\
Randomnodeedgemask2\_base & Hisat2, Scallop \\
Randomnodeedgemaskalign\_base & Samtools, Scallop \\
Lowerattention\_base & NCBI, Hisat2, Scallop \\
Higherattention\_base & NCBI, Samtools, Gffcomapre \\
Randomnodesmask1\_adapter & Hisat2, Scallop, Gffcompare, custom script \\
Randomnodesmask2\_adapter & NCBI, Hisat2, Scallop, GTF Annotation, Gffcompare \\
Randomnodesmaskalign\_adapter & Hisat2, Scallop, Gffcompare, custom script \\
Randomedgemask1\_adapter & NCBI, FastQC, Trimmonmatic, HISAT2, Samtools, Stringtie, GATK, VEP \\
Randomedgemask2\_adapter & Hisat2, Samtools, Stringtie, Cufflinks, Gffcompare \\
Randomnodeedgemask1\_adapter & Hisat2, Scallop, Gffcompare \\
Randomnodeedgemask2\_adapter & Trimmomatic, TrimGalaxy, Hisat2, Scallop, Stringtie, BLAST, T-coffee \\
Randomnodeedgemaskalign\_adapter & Samtools, Scallop, STAR, Hisat2, TransABySS, VarScan, Ensembl API \\
Lowerattention\_adapter & NCBI, Hisat2, Scallop \\
Higherattention\_adapter & NCBI, Hisat2, STAR, StringTie \\
\bottomrule
\end{tabular}
\caption{Extracted toolchains from model outputs under different structural perturbation settings.}
\label{tab:toolchains}
\end{table}

Table~\ref{tab:toolchains} presents the extracted toolchains from the results obtained in different baseline experiments, where various graph constructions (Figure~\ref{fig:all_masks}) were used as prompt templates in different LLMs. It is important to note that the presence of the NCBI database in the table indicates that, during the construction process, this node is retained, and the LLM generates new nodes around it. Given its significant role in the graph, it is preserved in the toolchain. Additionally, the position of NCBI in the graph allows for a quick assessment of the completeness of the toolchain.

\end{document}